%% file: main.tex
\documentclass[11pt,a4paper]{article}

\usepackage[utf8]{inputenc}
\usepackage[T1]{fontenc}
\usepackage{amsmath,amssymb,amsthm}
\usepackage{graphicx}
\usepackage{hyperref}
\usepackage{cleveref}
\usepackage{booktabs}
\usepackage{multirow}
\usepackage{xcolor}
\usepackage{enumitem}
\usepackage{geometry}
\usepackage{natbib}
\usepackage{caption}
\usepackage{subcaption}
\usepackage{tikz}
\usepackage{array}
\usepackage{tabularx}
\usepackage{float}
\usepackage{microtype}

\hypersetup{
    colorlinks=true,
    linkcolor=blue,
    citecolor=blue,
    urlcolor=blue
}

\newcommand{\tw}{\mathrel{\hspace{-0.5em}\wedge\hspace{-0.5em}}}

\title{\textbf{Geometric Reasoning in the Embedding Space}}

\author{
David Mojžíšek$^{1,*}$ \and 
Jan Hůla$^{2}$ \and 
Jiří Janeček$^{1}$ \and 
David Herel$^{2}$ \and 
Mikoláš Janota$^{2}$
\\[1em]
\small $^{1}$ University of Ostrava, Ostrava, Czechia \\
\small $^{2}$ Czech Technical University in Prague, Prague, Czechia \\[0.5em]
\small $^{*}$ Correspondence: david.mojzisek@osu.cz
}

\date{}

\begin{document}

\maketitle

\begin{center}
\fbox{\parbox{0.9\textwidth}{
\small\textbf{Published Version:} This paper has been published in \textit{Machine Learning and Knowledge Extraction} (MAKE), 2025, 7(3): 93. \\
\textbf{DOI:} \url{https://doi.org/10.3390/make7030093} \\
\textbf{Citation:} Mojžíšek, D.; Hůla, J.; Janeček, J.; Herel, D.; Janota, M. Geometric Reasoning in the Embedding Space. \textit{Mach. Learn. Knowl. Extr.} \textbf{2025}, \textit{7}(3), 93.
}}
\end{center}

\vspace{1em}

\begin{abstract}
While neural networks can solve complex geometric problems, as demonstrated by systems like AlphaGeometry, we have limited understanding of how they internally represent and reason about spatial relationships. In this work, we investigate how neural networks develop internal spatial understanding by training Graph Neural Networks and Transformers to predict point positions on a discrete 2D grid from geometric constraints that describe hidden figures. We show that both models develop interpretable internal representations that mirror the geometric structure of the problems they solve. Specifically, we observe that point embeddings self-organize into 2D grid structures during training, and during inference, the models iteratively construct the hidden geometric figures within their embedding spaces. Our analysis reveals how reasoning complexity correlates with prediction accuracy, and shows that models solve constraints through an iterative refinement process, which might resemble continuous optimization. We also find that Graph Neural Networks prove more suitable than Transformers for this type of structured constraint reasoning and scale more effectively to larger problems. These findings provide initial insights into how neural networks can develop structured understanding and contribute to interpretability of neural networks.
\end{abstract}


\vspace{1em}

\section{Introduction}
\label{sec:intro}
\input{sections/intro}

\section{Related Work}
\label{sec:related}
\input{sections/related_work}

\section{Problems and Methods}
\label{sec:setup}
\input{sections/experimental_setup}

\section{Experiments and Results}
\label{sec:results}
\input{sections/results}

\section{Limitations and Discussion}
\label{sec:limitations}
\input{sections/limitations}
\input{sections/fuwork}

\section{Conclusions}
\label{sec:conclusion}
\input{sections/conclusion}

\section*{Author Contributions}
Conceptualization, J.H. and D.M.; methodology, J.H. and D.M.; software, J.H., D.M., D.H. and M.J.; validation, D.M. and J.H.; formal analysis, D.M. and J.J.; investigation, D.M. and J.H.; resources, J.H. and M.J.; data curation, D.M.; writing---original draft preparation, D.M. and J.H.; writing---review and editing, J.J., D.M. and J.H.; visualization, D.M.; supervision, J.H. and M.J.; project administration, J.H. All authors have read and agreed to the published version of the manuscript.

\section*{Funding}
This article has been produced with the financial support of the European Union under the REFRESH -- Research Excellence For REgion Sustainability and High-tech Industries project number CZ.10.03.01/00/22\_003/0000048 via the Operational Programme Just Transition.

\section*{Data Availability}
The synthetic datasets and problem generator code supporting the findings of this study are available on request from the corresponding author. The codebase including model implementations and training procedures will be made publicly available after publication.

\section*{Acknowledgments}
During the preparation of this manuscript, the authors used Claude (Anthropic's AI assistant) for code development assistance and English language editing suggestions. All conceptualization, experimental design, analysis, and interpretation of results were conducted by the authors. The AI tool was used primarily for programming assistance and language refinement suggestions. The authors have reviewed and edited all output and take full responsibility for the content of this publication.

\section*{Conflicts of Interest}
The authors declare no conflicts of interest.

\section*{Abbreviations}
The following abbreviations are used in this manuscript:
\\[0.5em]
\begin{tabular}{@{}ll}
NN & Neural network\\
GNN & Graph neural network\\
CSP & Constraint satisfaction problem\\
SDP & Semidefinite programming\\
LLM & Large language model\\
RNN & Recurrent neural network\\
M & Midpoint constraint type\\
R & Reflection constraint type\\
S & Square constraint type\\
T & Translation constraint type\\
DAG & Directed acyclic graph\\
LSTM & Long short-term memory\\
CoT & Chain-of-thought\\
PCA & Principal component analysis\\
UMAP & Uniform manifold approximation and projection
\end{tabular}

\appendix
\label{sec:appendix}
\input{sections/appendix}

\bibliographystyle{plainnat}
\bibliography{references}

\end{document}

%% file: sections/intro.tex
While neural networks can solve complex spatial reasoning problems, we have limited understanding of the internal mechanisms they use to represent and manipulate geometric relationships. Many papers already demonstrated that autoregressive \emph{Transformers}, for example \cite{Momennejad2023EvaluatingCM,wu2024mind,Yamada2023EvaluatingSU}, and \emph{Graph Neural Networks} (GNNs), for example \cite{li2023depwignn, teodorescu2022spatialsim}, are able to learn to solve such problems but it is not clear what mechanism they discovered. 

A notable example is the work by \cite{trinh2024solving} who developed a system named \emph{AlphaGeometry}, which was able to solve problems that appeared on the International Mathematical Olympiad. The system contains an autoregressive Transformer that takes as input a sequence of tokens describing the problem in the language of geometric relations and is trained to predict auxiliary points useful for finding a proof for the given statement. AlphaGeometry achieves impressive performance, it operates as a black box, providing no insights into how the model internally represents geometric relationships or constructs spatial understanding.

A human dealing with such a problem would try to form a mental image of the construction used in the problem (most likely by first drawing it). We can ask a natural question whether NNs could also form such ``mental image'', which would reflect the spatial configuration of points described in the problem. We can also ask whether an autoregressive Transformer is a suitable model for such a problem and whether a GNN, which eliminates a lot of symmetries~\footnote{Variable renaming and constraint reordering.}, could be easier to train and be more scalable.

In this contribution, we take a closer look at these questions using a simplified approach. We will focus on purpose-built geometric \emph{Constraint Satisfaction Problems} (CSPs). This setting allows for detailed analysis of neural geometric reasoning that is difficult to achieve in complex systems like AlphaGeometry. We create a simple CSP language with several geometric constraints (relations) for which the domain is a set of points in a discrete 2D grid of certain size. Each instance of this CSP uniquely describes a hidden figure whose points are the solution of the instance. As the discrete grid is finite, we can assign a token/class~\footnote{When using a Transformer, each point is represented by a token; when using GNN, each point corresponds to a class.} to each point and train the model to predict the points in the hidden figure with a cross-entropy loss. 

Analysis shows that both models develop interpretable internal representations that mirror the geometric structure of the problems they solve. After visualizing a low-dimensional projection of the embeddings corresponding to individual points, we can see that they organized themselves in a 2D grid which they represent. We also show that during inference, the embeddings of unknown variables evolve into a configuration that reflects the hidden figure described by the problem. Additionally, we find that GNNs prove more suitable than Transformers for structured constraint reasoning tasks and scale more effectively to larger problems.

This controlled approach provides solid mechanistic insights into how neural networks develop spatial understanding and contributes to broader understanding of structured reasoning in neural systems.

The rest of the paper has the following structure. In Section~\ref{sec:related}, we review related work. Section~\ref{sec:setup} describes our experimental setup, including the two types of architectures and the process for generating problems. In Section~\ref{sec:results}, we present results of experiments for both models, analyze embedding structure emergence, examine the iterative solution process, and provide analysis of failure modes for different problem complexities. We discuss the results and limitations of our work in Section~\ref{sec:limitations}. Supplementary materials can be found in Appendix.

%% file: sections/related_work.tex
\subsection{Reasoning About Geometry}
Geometric reasoning has been a focus of research for decades \cite{wen1986basic}, with Wu's method \cite{wu2008decision} considered state-of-the-art until recently. In 2024, AlphaGeometry \cite{trinh2024solving} surpassed Wu's method on International Mathematical Olympiad problems, utilizing a combination of symbolic deduction and neural language model which predicts auxiliary constructions based on problem statement which is described using geometric relations. Our work is partially inspired by AlphaGeometry but we study much easier setup in which the model predicts points in a 2D grid that satisfy the relations used to describe the construction. 

Other neural approaches include visual reasoning methods that extract geometric primitives from diagrams using deep learning for construction problems \cite{wong2022euclidnet}. Our work differs by focusing on abstract constraint satisfaction rather than visual diagram processing, enabling detailed analysis of internal spatial representations.

We were also partially inspired by the work of \cite{hula2024revisiting} which provides explanation for the process by which GNNs can learn to solve Boolean CSPs, concretely deciding satisfiability of Boolean formulas. We augment their setup for the domain of geometric CSP. In their work, they show an empirical evidence that the GNN learns to act as a first-order solver of a \emph{semidefinite programming} (SDP) relaxation of MAX-SAT. MAX-SAT is an optimization version of SAT and in the SDP relaxation, vectors assigned to variables are being optimized in order to minimize an objective which captures how many constraints are satisfied. 

In the domain of geometric CSPs, \cite{krueger2021automatically} developed an algorithm for automatic construction of diagrams for geometry problems. The algorithm works by optimizing coordinates of points by gradient descent to maximize an objective function which captures how well are the geometric constraints of the problem satisfied. We emphasize that their algorithm was manually constructed and not learned.

\subsection{Spatial Reasoning in Neural Networks}
Another related area of research is focused on the ability of NNs to do spatial reasoning in a broader sense, for example spatial navigation, planning and interaction with physical objects. 

Several papers explore spatial reasoning in the context of Large Language Models (LLMs) such as \emph{ChatGPT} or \emph{LLama}  \cite{touvron2023llama}. In \cite{Momennejad2023EvaluatingCM} and \cite{Yamada2023EvaluatingSU}, they point out various failure modes in a planning and navigation capability of large pool of LLMs. In \cite{wu2024mind} they developed a prompting technique called Visualization-of-Thought (VoT), which significantly improved LLM performance in visual navigation tasks. Most similar to our work is the work by \cite{Ivanitskiy2023StructuredWR} who probe small Transformer models trained for maze navigation in order to understand the representation it uses to solve the task. Unlike the previously mentioned works, \cite{janner2018representation} uses reinforcement learning to train a language model based on a recurrent NN (RNN)
for text-based spatial reasoning. 

Apart from language models, other papers also explore spatial reasoning in the context of GNNs. In \cite{li2023depwignn} they developed a GNN for spatial navigation, 
which utilizes a memory mechanism for long-range information storage and \cite{teodorescu2022spatialsim} introduced the \emph{SpatialSim} benchmark together with a GNN that is trained to recognize spatial configurations. Our work contributes into this area by partially elucidating the mechanism by which NNs are able to reason about geometric relations and by showing that GNNs which use the graph of relations/constraints and variables are much easier to train and scale.

%% file: sections/experimental_setup.tex
To test the ability to reason about geometric relations, we develop a simple generative procedure of problems with solutions that allows to control the difficulty and size of the problem. We test two types of models, GNNs and Transformers, on data generated with various generator settings.

\subsection{Geometric Problem Generation}
\label{sec:data_gen}
Our goal was to create a simple synthetic dataset that would enable us to study the mechanism which NNs use to reason about geometric relations. Specifically, we wanted to investigate whether neural networks develop internal representations that reflect the geometric structure of the problems they solve. In the task of auxiliary point prediction for which AlphaGeometry was using a language model, it is, for example, not clear whether the model is taking the geometric nature of the constraints into account or treats the language of the problem abstractly without any metric interpretation/grounding. 

Therefore, we designed a generator for CSPs whose solutions can be interpreted as geometric figures described by the input constraints. The task of the model is to predict positions of the points contained in the constraints. We simplify the problem so that the geometric figures lie on a discrete grid, and we can therefore treat the prediction of the positions as a classification task. We note that one can also treat the task as a regression and train the model to process figures with arbitrary 2D coordinates.

Our CSP is using four types of constraints: $M$, $R$, $S$ and $T$. The constraint~$M(A, B, C)$ says that the point $B$ is a midpoint of the line segment given by $A$ and $C$ and $R(A, B, C, D)$ says that points $A$, $B$ give an axis of symmetry around which $C$ and $D$ reflect each other. The constraint~$S(A, B, C, D)$ says that points $A$, $B$, $C$, $D$ form a square and $T(A, B, C, D)$ says that the vector $D-C$ is a translation of the vector $B-A$. In order to have unique solutions, we also need to fix several points to concrete positions. We can achieve that by additional constraint $P$ for which $P(a, [1, 1])$ signifies that point $A$ lies at position $[1, 1]$. For each constraint, certain number of variables are required to be known so that the constraint can be uniquely resolved, here we call them \emph{determining} variables. Once these variables are fixed, the rest of variables is uniquely determined. These are called \emph{dependent} variables. For the constraint~$M$, $2$ variables are determining and $1$ is dependent, i.e. if we know positions of $2$ variables, the last one could be resolved. For the constraint~$R$, $3$ variables are determining and $1$ is dependent.  For the constraint~$S$, $2$ variables are determining and $2$ are dependent (i.e., we assume a fixed spatial ordering of the square vertices). For the constraint~$T$, $3$ variables are determining and $1$ is dependent. The last subplot in Figure \ref{fig:point_evo} shows a visualization of the following instance of the CSP: 
\begin{align*}
& T(F,D,E,G) && \tw && T(C,G,A,H) && \tw && S(H,I,G,J) && \tw && S(B,J,L,K) && \tw && P(A,[6,15]) && \tw \\
& P(B,[12,16]) && \tw && P(C,[12,14]) && \tw && P(D,[14,12]) && \tw && P(E,[12,14]) && \tw && P(F,[12,9]) &&\ \ \ \ \ ,
\end{align*}
where $A-F$ are known points and $G-L$ are unknown.

Our generator creates problems that require deductive reasoning by ensuring constraints form a dependency structure where some must be resolved before others. Attempting to solve all constraints simultaneously is not feasible because dependent variables would be incorrectly determined. For example, a network might successfully create a square from four unknown points, but if some of those points are actually determined by previous translation or reflection constraints, the resulting square would be geometrically valid but positioned incorrectly within the overall figure. This dependency structure forces models to discover the proper reasoning sequence rather than solving constraints in isolation.

To create a single problem, we sample a random \emph{directed acyclic graph} (DAG) with nodes corresponding to types of constraints and edges corresponding to dependencies. Then we add variables to the constraints so that each constraint within graph can be uniquely resolved given that the parent constraints are already resolved (the parent constraints therefore need to contain the determining variables). To produce a problem with a unique solution, we fix the required number of variables appearing in each of the root constraints which have no parents (i.e. arbitrary two variables in the S constraint, three variables for the T constraint, etc.) We use the \emph{Z3} solver \cite{de2008z3} to obtain the assignment for all other points. If the resulting figure is larger than the size of the grid, we reject the problem, otherwise we place it to a random position within the grid. The constraints with the fixed points are given as an input to the model, the positions of the remaining variables are given as labels.

By varying the number and types of constraints, the generator produces problems with different complexity levels and reasoning depths. This allows us to investigate whether models can solve problems harder than those seen during training by leveraging test-time scaling approaches such as increased iterations or resampling.

\subsection{Models}
In our experiments, we tested two types of architectures: GNNs and autoregressive Transformers.

\paragraph{Graph Neural Network}
The architecture of the GNN is inspired by the model presented in \cite{hula2024revisiting} and originated in \cite{selsam2018learning}. Similarly as in works about Boolean satisfiability, it applies the same update rules repeatedly. It is a GNN operating on a bipartite graph with $n$ variables and $m$ constraints defined by Equations \eqref{eq:X_update} and \eqref{eq:Z_update} below which are applied recurrently for several rounds. Equation~\eqref{eq:X_update} updates the embeddings of variables which are, after the last update, used to classify a given variable to a point within the grid. This is achieved by applying a linear layer $L(\cdot)$ on each variable embedding. As mentioned in Section \ref{sec:data_gen}, some variables need to be fixed to concrete points so that the problem has a unique solution. 

The embeddings of known variables are initialized using the embedding layer that shares weights with the final classification layer $L(\cdot)$, i.e., both use the same, trainable, weight matrix $\mathbf{W} \in \mathbb{R}^{N \times d}$, where $N$ is the number of grid positions and $d$ is the embedding dimension. This design ensures consistent representation of grid positions throughout the network.\footnote{Similarly as token embeddings share weights with the classification head in language models.} They are not updated during message passing. The embeddings of unknown variables are initialized randomly, i.e., they are sampled from a unit hypercube $[0, 1]^d$ following the uniform probability distribution of each coordinate. It enables resampling during inference by using different random initializations.

The embeddings are updated after each message passing iteration. Let $n$ be the number of variables, $m$ be the number of constraints. The update equations have the following form:

\begin{align}
Z^{t+1}_c &= U_{c}\left(\text{cat}\left(A_{c} X^t\right), Z^t_c, \Phi_{c}\right) &  \forall c \in C\tag{1}\label{eq:Z_update}\,, \\
X^{t+1} &= U_{X}\left(A_{X} Z^{t+1}, X^t, \Phi_X\right) \tag{2}\label{eq:X_update}\,,
\end{align}

where $X^t \in \mathbb{R}^{n \times d}$ is the matrix of stacked variable embeddings, $Z^t \in \mathbb{R}^{m \times d}$ is the matrix of stacked constraint embeddings, $A_X \in \mathbb{R}^{n \times m}$ is the adjacency matrix that matches constraints to variables, $A_c \in \mathbb{R}^{4m \times n}$ is the adjacency matrix that selects variables for each constraint\footnote{$A_{c} X^t$ will have $4m$ rows because there are 4 variables in each constraint. For constraints that have only three variables, we append a zero vector to preserve the shape.} (taking the constraint type and argument position into account), and $\text{cat}(A_c X^t) \in \mathbb{R}^{m \times 4d}$ denotes the operation that extracts and concatenates the embeddings of the four variables involved in each constraint in the correct order. $C = \{M, R, S, T\}$ is the set of constraint types. Each problem instance is represented by a bipartite graph containing the denotation of the constraint types, which is fully encoded by the matrix $A_c$, from which the matrix $A_X$ can be easily determined.

\begin{figure*}[ht]
    \centering
    \includegraphics[width=0.9\textwidth]{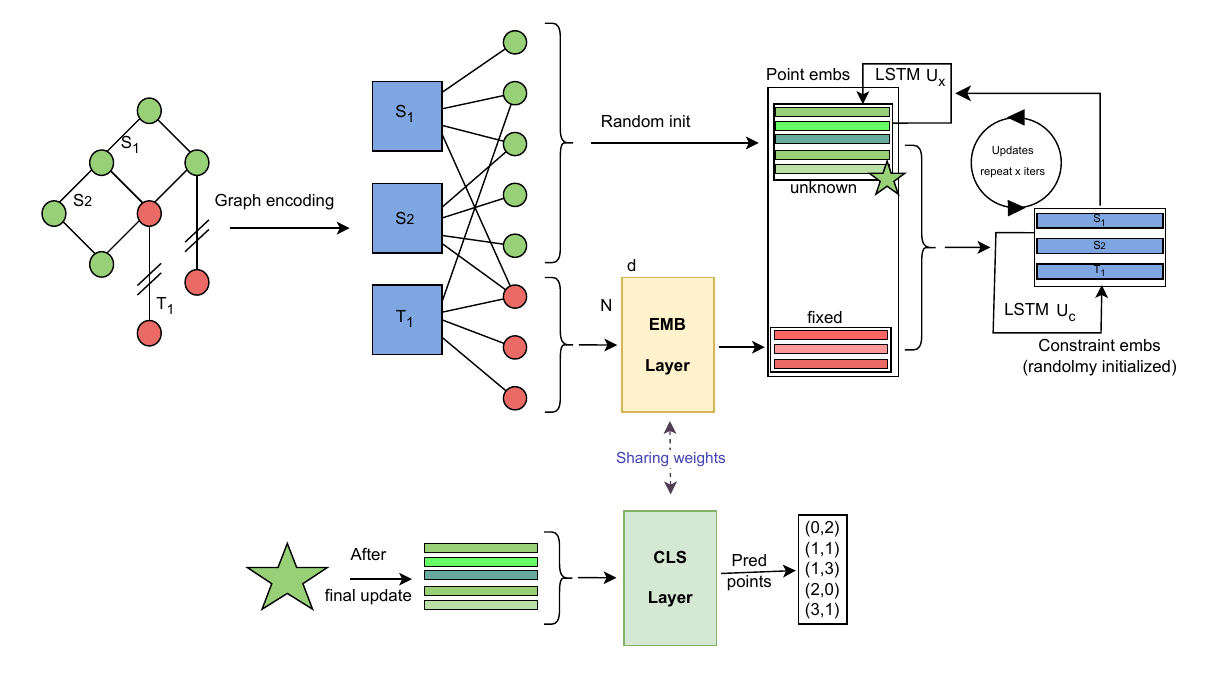}
    \caption{Solution process of the GNN model. The geometric problem is encoded as a bipartite graph with variable nodes (points) and constraint nodes. Red points represent fixed/known variables that maintain their initial embeddings from the embedding layer throughout the process. Green points represent unknown variables and blue nodes represent constraint embeddings, both initialized randomly. The embedding layer is represented as a matrix with dimensions $N$ (corresponding to grid points i.e. classes) by $d$ (embedding dimension, $d=128$). During inference, unknown point embeddings and constraint embeddings are iteratively updated using respective LSTMs: $U_x$ updates variable embeddings (Equation \ref{eq:X_update}) and $U_c$ updates constraint embeddings (Equation \ref{eq:Z_update}) for a specified number of iterations. After the final update, unknown points are classified using a linear classification layer without bias that shares weights with the embedding layer. The classifier produces logits that are converted to probabilities via softmax to predict the final point positions.}
    \label{fig:diagf}
\end{figure*}

Equation~\eqref{eq:X_update} updates the embeddings of variables and Equation~\eqref{eq:Z_update} updates the embeddings of the constraints. The latter equation is further indexed by $c$ which reflects the fact that we are using multiple types of constraints which require distinct update functions. The update functions $U_x$ and $U_c$ ($\forall c \in C$)  are in our case realized as \emph{LSTMs} \citep{hochreiter1997long} parameterized by $\Phi_X$ and $\Phi_c$, respectively. Technically, each LSTM also updates a cell state which is not mentioned in the equations. These LSTMs could be replaced by simple RNNs but we found the LSTMs easier to optimize and RNN does not work well with our data, see \ref{app:rnn_ablation}. The simplified schema of the GNN processing an example can be found in Figure \ref{fig:diagf}.

The update of each embedding happens independently of the other embeddings, i.e. in Equation~\eqref{eq:X_update}, each embedding of a variable (row in the matrix $X^t$) is updated independently by the same LSTM which takes as input the aggregated message from the constraints containing this particular variable. The aggregated message is simply the sum of the relevant constraint embeddings and it is realized by multiplying the constraint embedding matrix $Z^t$ by the matrix $A_X$. The LSTM $U_c$ which updates the constraint embeddings, differs by the fact that the aggregated message is not obtained as a sum but as a concatenation of variable embeddings. The embeddings are concatenated in the order in which the variables appeared within the constraints. For example, for the constraint $S(A, B, C, D)$, we concatenate the embeddings of variables $A,B,C,D$ in that order. One can intuitively view the embeddings of variables as if they are representing the values of these variables (points) and embeddings of constraints as if they represent the information of what is needed to satisfy the constraint. For this reason, the function which updates the constraint embeddings cannot be permutation invariant because different order of the determining variables results in different values for the dependent variables. 

The embedding dimension $d$ and number of iterations are configurable parameters that we adjust based on problem complexity. Importantly, since the same update rules are applied recurrently, the number of iterations can be modified at test time without changing the model parameters. This architectural property, leads to the option of allocating more computation to solve harder problems during inference. Other hyperparameters for training the GNN can be found in Table~\ref{tab:hyperparams}. 

For clarity, one illustrative example of how a GNN processes and solves a specific problem will be presented later in \ref{sec:detailed_example}.

\paragraph{Autoregressive Transformer}
The autoregressive Transformer model is based on the GPT-2 architecture developed by \cite{radford2019language} with rotary embeddings \cite{su2024roformer}. It takes as input a sequence of tokens representing the problem together with a query for a variable we want to predict. For example, the input with just one constraint could have the following form: S ( [0,0] [0,1] C D ) ? D. It is querying the position for variable $D$ within a square with two known points. The model reads this sequence of tokens (where the positions of points such as $[0,0]$ correspond to one token) and is trained to predict the token which corresponds to the the correct position of variable $D$ ($[1,0]$ in this case). This means that from each CSP produced by the generator described in Section \ref{sec:data_gen}, we extract one sequence for each unknown point.  
After experimenting with the hyperparameters of the model, we set the number of layers to 6, number of heads to 6 and the embedding dimension to 256. We also experimented with a recurrent application of one layer, as done in \cite{dehghani2018universal}, to more closely mimic the GNN, but having separate weights for each layer produced better results. 
Other hyperparameters of the model and for training can be found in Table~\ref{tab:hyperparams}.

\subsection{GNN: Illustrative Example}
\label{sec:detailed_example}

\begin{figure*}[ht]
    \centering
    \includegraphics[width=0.9\textwidth]{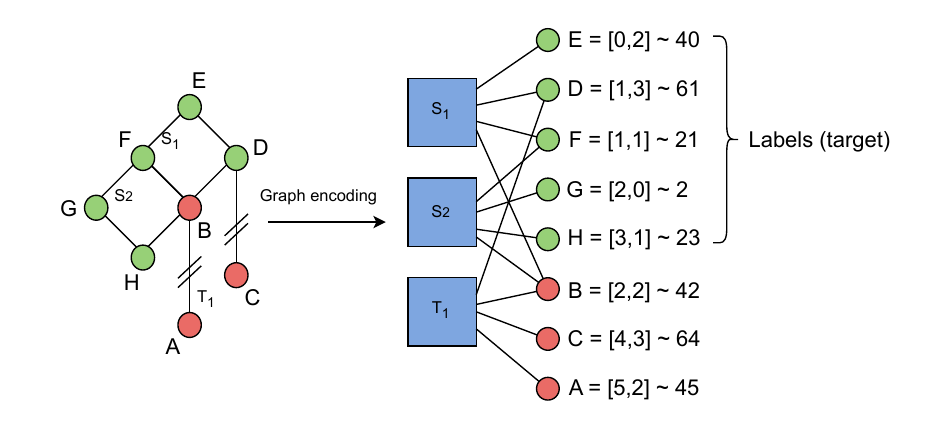}
    \caption{Graph encoding of the illustrative example problem showing the bipartite structure with point nodes and constraint nodes, along with the point-to-index mapping for the 20×20 grid.}
    \label{fig:ilexample}
\end{figure*}

To illustrate how the GNN works, we present a concrete example showing how it processes geometric constraints on a $N=20\times20$ grid with embedding dimension $d = 128$ and iteration count $t=15$. Consider the following constraint satisfaction problem:

\begin{align*}
&T(A, B, C, D) \land S(B, D, E, F) \land S(B, F, G, H)\land\\
&P(A, [5, 2]) \land P(B, [2, 2]) \land P(C, [4, 3])
\end{align*}

where $A, B, C$ are known points fixed by position constraints $P$, and $D, E, F, G, H$ are unknown points to be predicted.

\subsubsection*{Problem Encoding and Dependencies}

Each grid position $(x, y)$ maps to a unique index using $\text{index} = x + y \times 20$. Therefore, the known points correspond to indices: $A \rightarrow 45$, $B \rightarrow 42$, $C \rightarrow 64$. The problem forms a bipartite graph with 8 point nodes and 3 constraint nodes, where the dependency structure requires sequential resolution: constraint $T_1$ determines point $D$, which enables constraint $S_1$ to determine points $E$ and $F$, which finally allows constraint $S_2$ to determine points $G$ and $H$. 

The point constraints $P$ are not realized as separate constraint nodes in the bipartite graph, but rather as fixed embeddings taken directly from the embedding layer. Figure \ref{fig:ilexample} illustrates this encoding and the target solution.

\subsubsection*{Embedding Initialization}

The model uses a shared embedding matrix $\mathbf{W} \in \mathbb{R}^{400 \times 128}$ for both the embedding layer and final classifier. Known points are initialized with fixed embeddings $\mathbf{W}_{45}$, $\mathbf{W}_{42}$, $\mathbf{W}_{64} \in \mathbb{R}^{128}$ corresponding to points $A$, $B$, $C$ respectively, where $\mathbf{W}_{i}$ denotes the $i$-th row of matrix $\mathbf{W}$. These embeddings remain unchanged throughout inference: \[
\mathbf{x}_A^{(t)} = \mathbf{W}_{45}, \qquad
\mathbf{x}_B^{(t)} = \mathbf{W}_{42}, \qquad
\mathbf{x}_C^{(t)} = \mathbf{W}_{64}
\quad \text{for all } t.
\]

Unknown points are initialized with random vectors $\mathbf{r}_D, \mathbf{r}_E, \mathbf{r}_F, \mathbf{r}_G, \mathbf{r}_H \sim \mathcal{U}(0, 1)^{128}$, sampled uniformly from the unit hypercube, and constraint embeddings are initialized as random vectors $\mathbf{c}_1, \mathbf{c}_2, \mathbf{c}_3 \sim \mathcal{U}(0, 1)^{128}$. These random vectors are used as the initial hidden states for the LSTM-based message passing. Specifically, for each unknown point $i$, we set the initial embedding as the hidden state $\mathbf{x}_i^{(0)} := \mathbf{r}_i$, and similarly for each constraint $c$, we set $\mathbf{z}_c^{(0)} := \mathbf{c}_c$. The LSTM cell states $\mathbf{h}_i^{(0)}$ and $\mathbf{h}_c^{(0)}$ are initialized to zero. During inference, the hidden states $\mathbf{x}_i^{(t)}$ and $\mathbf{z}_c^{(t)}$ evolve over time according to the update equations.

\subsubsection*{Message Passing Dynamics}

The update equations are repeated for 15 iterations. Our LSTM-based architecture maintains both hidden states and cell states, where the \textit{hidden states serve as the embeddings} that flow through the message passing network, while cell states maintain internal memory. The LSTM also produces a cell output at each iteration, but this output is discarded and only the hidden and cell states are retained.

For constraint updates (Equation \ref{eq:Z_update}), each constraint type uses a specialized LSTM update function. The constraint embedding $\mathbf{z}_c^{(t)}$ (LSTM hidden state) and cell state $\mathbf{h}_c^{(t)}$ are updated as:

$$\mathbf{z}_c^{(t+1)}, \mathbf{h}_c^{(t+1)} = U_c^{(\text{type})}(\mathbf{m}_c^{(t)}, \mathbf{z}_c^{(t)}, \mathbf{h}_c^{(t)})$$

where $\mathbf{m}_c^{(t)}$ is the concatenated message from variables participating in constraint $c$, and each constraint type has its own LSTM parameters: $U_c^{(T)}$ for translation, $U_c^{(S)}$ for square, $U_c^{(R)}$ for reflection, and $U_c^{(M)}$ for midpoint constraints. The LSTM returns both an output (which is discarded) and a tuple containing the new hidden state $\mathbf{z}_c^{(t+1)}$ and cell state $\mathbf{h}_c^{(t+1)}$.

For the above example, the input message $\mathbf{m}_{T_1}^{(t)}$ for constraint $T_1$ is the concatenated vector $[\mathbf{W}_{45}; \mathbf{W}_{42}; \mathbf{W}_{64}; \mathbf{x}_D^{(t)}] \in \mathbb{R}^{512}$, where $\mathbf{W}_{45}$, $\mathbf{W}_{42}$, and $\mathbf{W}_{64}$ are the fixed embeddings of known points $A$, $B$, and $C$ respectively, and $\mathbf{x}_D^{(t)}$ is the current hidden state of point $D$. Similarly, the messages $\mathbf{m}_{S_1}^{(t)}$ and $\mathbf{m}_{S_2}^{(t)}$ for constraints $S_1$ and $S_2$ are given by $[\mathbf{W}_{42}; \mathbf{x}_D^{(t)}; \mathbf{x}_E^{(t)}; \mathbf{x}_F^{(t)}]$ and $[\mathbf{W}_{42}; \mathbf{x}_F^{(t)}; \mathbf{x}_G^{(t)}; \mathbf{x}_H^{(t)}]$ respectively. Each of these message vectors is passed to the corresponding constraint-specific LSTM update function $U_c^{(\text{type})}$.

For variable updates (Equation~\ref{eq:X_update}), unknown point embeddings are updated using messages aggregated from all connected constraints. Each unknown variable $i$ aggregates the sum of constraint embeddings from all connected constraints and passes this sum to the shared LSTM update function $U_X$:

\[
\mathbf{x}_i^{(t+1)}, \mathbf{h}_i^{(t+1)} = U_X\left(\sum_{c \in \mathcal{N}(i)} \mathbf{z}_c^{(t+1)}, \mathbf{x}_i^{(t)}, \mathbf{h}_i^{(t)}\right)
\]

Here, $\mathcal{N}(i)$ denotes the set of constraint nodes connected to variable $i$, $\mathbf{x}_i^{(t)}$ is the hidden state (embedding) of point $i$ at time step $t$, and $\mathbf{h}_i^{(t)}$ is its corresponding LSTM cell state.

In the illustrative example, point $D$ receives messages from constraints $T_1$ and $S_1$, resulting in an update input of $\mathbf{z}_{T_1}^{(t+1)} + \mathbf{z}_{S_1}^{(t+1)}$. Point $F$ receives messages from $S_1$ and $S_2$. Points $E$, $G$, and $H$ each receive a single constraint message: from $S_1$ for $E$, and from $S_2$ for $G$ and $H$.

These aggregated messages are passed into the $U_X$ LSTM, which outputs the updated hidden state $\mathbf{x}_i^{(t+1)}$ that becomes the new embedding of variable $i$.

\subsubsection*{Solution Recovery}

After $t$ iterations, the model predicts the final grid position of each unknown variable using a standard classification head. The final embedding $\mathbf{x}_i^{(T)}$ is projected using the shared embedding matrix $\mathbf{W} \in \mathbb{R}^{400 \times 128}$ (no bias term) to produce logits:
\[
\boldsymbol{\ell}_i = \mathbf{W} \mathbf{x}_i^{(T)} \in \mathbb{R}^{400}
\]
The predicted index is obtained as $\hat{k}_i = \arg\max(\boldsymbol{\ell}_i)$, which is then mapped back to coordinates via $\hat{x} = \hat{k}_i \bmod 20$, $\hat{y} = \lfloor \hat{k}_i / 20 \rfloor$.

During training, we apply a cross-entropy loss over the logits $\boldsymbol{\ell}_i$, comparing predicted and ground-truth indices for all unknown variables. The embedding matrix $\mathbf{W}$ is shared with the embedding layer used for known variables, ensuring consistency between input encoding and prediction.

In our example, the correct predictions for points $D$–$H$ are $[1,3]$, $[0,2]$, $[1,1]$, $[2,0]$, and $[3,1]$, satisfying all constraints.

%% file: sections/results.tex
We evaluate both architectures on geometric constraint problems. We examine prediction accuracy, dataset complexity effects, embedding structure emergence, initialization strategies, solution dynamics, and failure modes.

\subsection{Comparison of the Two Architectures}\label{subsComparison} 
To compare the two architectures, we measure the accuracy of individual point prediction (point accuracy). In later experiments, we also report the accuracy in terms of correctly assigning all variables within the problem (complete accuracy).

The experiments on the synthetically generated data show that the GNN performs significantly better than the Transformer, as expected. In Appendix 
\ref{sec:scaling_laws}, we show that we were able to train the GNN on grid sizes up to $80 \times 80$ points to a validation accuracy larger than $90\,\%$. In comparison, the Transformer achieved accuracy of $90\,\%$ only if we train it on a grid size $10 \times 10$ and limit the number of constraints to $2$. If we train the model on a more complex setting with the grid size $20 \times 20$ and up to $6$ constraints, it reaches an accuracy of approximately $30\,\%$.

For completeness, we also trained the Transformer model to find the assignments to variables using \emph{Chain-of-Thought} (CoT) \cite{wei2022chain} by imitating a log from a simple solver. The solver resolves the constraints one by one in the topological order of the DAG of constraints mentioned in Section \ref{sec:data_gen}. Using this way of training, the Transformer learns to predict the positions of variables within $20 \times 20$ grid with approximately 50\% point accuracy. We did not explore this direction further as reasoning with a CoT is orthogonal to reasoning in the embedding space on which we focus in this work. More details about the CoT experiment can be found in Appendix~\ref{AsecCoT}.

Given the substantially better results achieved by the GNN on this problem setting, we conduct the main analysis of the embeddings with the GNN. Appendix~\ref{AsecC} 
contains similar analysis and visualization for the Transformer (see Figure~\ref{fig:tformer_grid}).

 We also note that the model accuracy can be further improved by leveraging multiple prediction attempts since incorrect predictions often fall close to their true positions (as shown in Appendix \ref{Adistanceerror}) and according to preliminary tests, sampling multiple different initial embeddings increases the chance of predicting the correct assignment. 

\subsection{GNN Performance Analysis}
\label{sec:gnn_performance}

Given the GNN's clear advantage over the Transformer for our CSPs representable by bipartite graphs, we now analyze the GNN's behavior in detail across multiple dimensions. We examine how the model performs under distribution shift, visualize the internal representations it develops, and identify failure modes related to problem complexity. The analysis uses our best-performing GNN configuration with embedding dimension $128$, $15$ message-passing iterations, and training procedures detailed in Appendix~\ref{APhyper}. This systematic evaluation reveals both the capabilities and limitations of our geometric constraint solving approach on discrete grids.

\subsubsection{Dataset Design and Complexity Distribution}

\begin{figure*}[ht]
    \centering
    \includegraphics[width=0.9\textwidth]{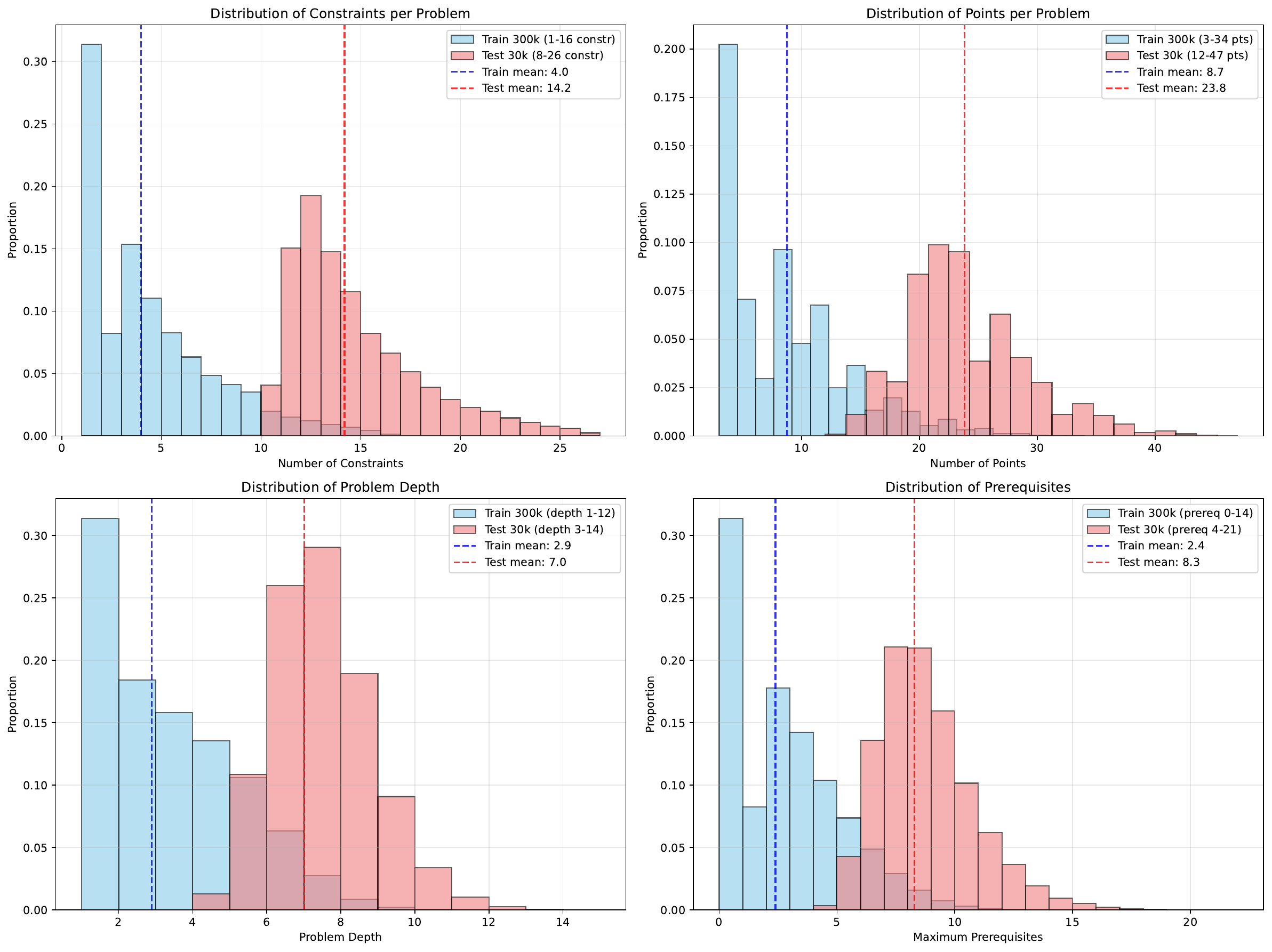}
    \caption{Deliberate distribution shift between training and test sets across complexity metrics. The harder test distribution enables analysis of model behavior under increased problem complexity, failure mode identification, and evaluation of test-time scaling approaches.}
    \label{fig:dataset_dist_comp}
\end{figure*}

The training dataset contains problems with $1$--$16$ geometric constraints (mean: $4.0$), $3$--$34$ points (mean: $8.7$), and reasoning depths of $1$--$12$ (mean: $2.9$). The test dataset was specifically designed to be more challenging, containing problems with $8$--$26$ geometric constraints (mean: $14.2$), $12$--$47$ points (mean: $23.8$), and reasoning depths of $3$--$14$ (mean: $7.0$). Figure~\ref{fig:dataset_dist_comp} illustrates these distribution differences across key complexity metrics.

\begin{figure}[ht]
    \centering
    \includegraphics[width=0.5\textwidth]{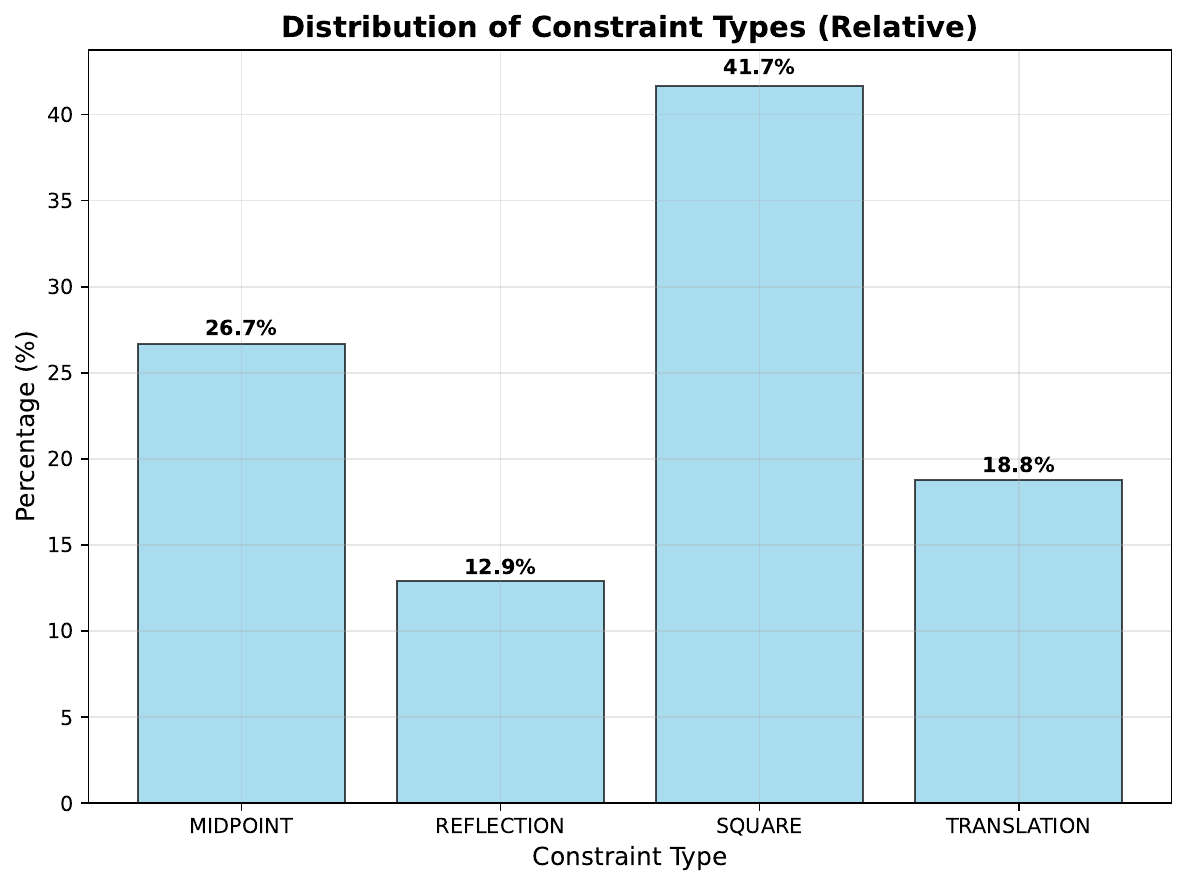}
    \caption{Distribution of constraint types in the training dataset. Square constraints are most frequent (41.7\%), followed by Midpoint (26.7~\%), Translation (18.8~\%), and Reflection (12.9~\%). The test dataset follows a very similar distribution, ensuring consistent constraint type representation across training and evaluation.}
    \label{fig:constraint_types}
\end{figure}

Our generator creates problems for specific grid sizes, and we focus on a $20 \times 20$ grid as it provides a good balance. It is large enough to accommodate interesting geometric figures while remaining computationally tractable for detailed analysis, which required multiple re-runs. Our constraint set consists of four types: Square (S), Reflection (R), Midpoint (M), and Translation (T), with their relative distribution in the training dataset shown in Figure~\ref{fig:constraint_types}.

This choice of difficulty distributions allows us to study model behavior when problem complexity is increased beyond what model encountered during training. The high validation accuracy achieved on the training distribution ($99.5$~\% point accuracy, $98.1$~\% complete problem accuracy using $10$\,~\% of training data as validation) demonstrates that the GNN can effectively solve geometric constraint problems when sufficient training data is available. We used $300 000$ ($270 000$ without validation) training examples. The size was determined through scaling analysis outlined in Appendix
\ref{sec:scaling_laws}, which shows the sample complexity requirements for different grid sizes.

The deliberately harder test distribution provides a challenging evaluation setting, where the model achieves lower baseline accuracy, creating opportunities to study failure modes, reasoning depth effects, and test-time scaling approaches.

\subsubsection{Performance under Distribution Shift}

Table~\ref{tab:spmd_results} presents comprehensive performance results on our datasets. The model maintains good performance even under distribution shift, achieving $96.07$~\% point accuracy and $76.74$~\% complete accuracy on the harder test set with standard inference ($15$ iterations, $1$ resample).

The results demonstrate clear benefits from test-time scaling approaches. Increasing inference iterations from $15$ to $23$ improves complete accuracy from $76.74$\,~\% to $93.62$\,~\%. Using $10$ resamples with different random initializations further boosts performance to $95.37$~\% complete accuracy. We determined the optimal iteration count of $23$ through analysis shown in Figure~\ref{fig:line1} which also reveals a slight performance increase with multiple resamples.

The ``Best'' configuration could be explained as an oracle that knows the optimal number of iterations for each individual problem to achieve the most solved points in the fewest iterations (with a maximum of $50$ iterations). This approach achieves the complete accuracy of $96.14$\,~\% with single resampling and $97.46$\,\% with $10$ resamples. This model could in theory get very close to it performance on validation set.

For resampling strategy, we tested various counts but found the most significant improvement occurs when going from $1$ to $5$ resamples. We chose $10$ resamples as a conservative buffer, since additional resamples beyond this point show diminishing returns while increasing computational cost.

These findings suggest several important properties of the learned model. The improvement from additional iterations indicates that the GNN employs an iterative refinement process similar to continuous optimization methods. The benefits of resampling show that different random initializations of unknown variables can lead to different solution paths, with some being more effective for particular problem instances.

Model's ability to generalize to problems with increased complexity beyond the training distribution, are consistent with previous work that applied a very similar GNN architecture to SAT problems \cite{mojvzivsek2025neural}.

\begin{table}
\centering
\caption{Performance of the model. The first row reports validation accuracy (on a separate dataset, using 1 resample). The remaining rows show results on a harder test set drawn from a different distribution, across different numbers of inference iterations (15, 23, Best) and resample counts (1 and 10). We report point-wise accuracy and complete accuracy. \vspace{0.5em}}
\vspace{10pt}
\label{tab:spmd_results}
\begin{tabular}{lcccc}
\toprule
 & \multicolumn{2}{c}{1 Resample} & \multicolumn{2}{c}{10 Resamples} \\
\cmidrule(lr){2-3} \cmidrule(lr){4-5}
Setting & Point Acc. & Complete Acc. & Point Acc. & Complete Acc. \\
\midrule
Validation (15 iters) & 99.55\,\% & 98.93\,\% & -- & -- \\
\midrule
Test (15 iters) & 96.07\,\% & 76.74\,\% & 97.89\,\% & 80.93\,\% \\
Test (23 iters) & 98.51\,\% & 93.62\,\% & 99.02\,\% & 95.37\,\% \\
Test (Best) & 99.04\,\% & 96.14\,\% & 99.44\,\% & 97.46\,\% \\
\bottomrule
\end{tabular}
\end{table}

\subsection{Visualizing the Embeddings of Individual Points}\label{sec:vis}
In the following text, we use the terms \emph{static embeddings} and \emph{dynamic embeddings}. By static embeddings we mean the embeddings of points of the grid which are used for initialization of known points and are shared with the classification head of both models. By dynamic embeddings we mean the embeddings of the unknown variables which are updated throughout the forward process. 

Both types of models are trained to predict positions of unknown points within the instance. This is done by a linear layer which computes the logits for each point. As already mentioned, the weights of this layer are shared with the embedding layer representing the position of known points. 

\begin{figure*}[ht]
    \centering
    \includegraphics[width=0.9\textwidth]{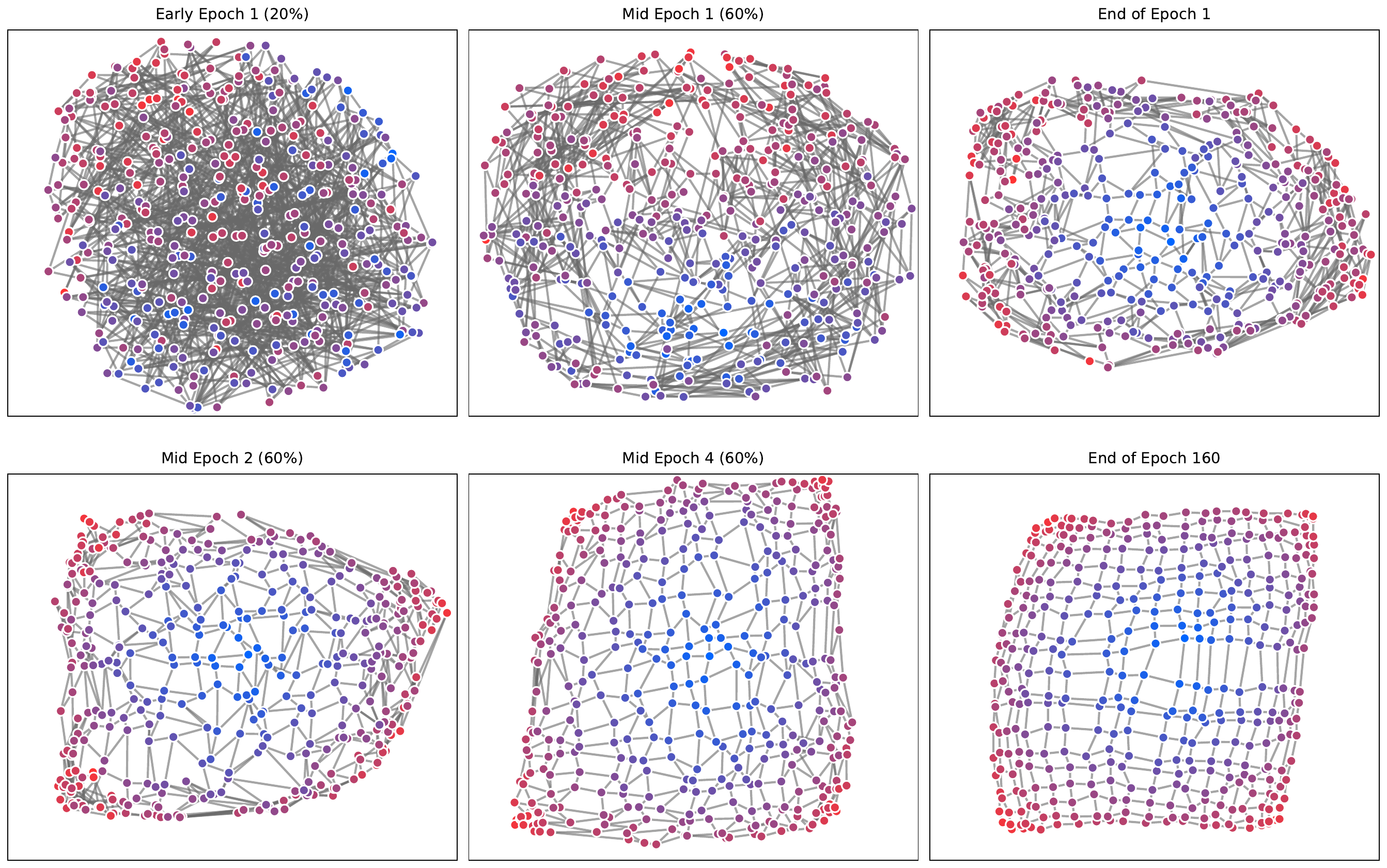}
    \caption{Emergence of spatial structure in static point embeddings during training on 20$\times$20 grid. Colors indicate distance from grid center, revealing how the model learns geometric relationships without explicit spatial supervision. UMAP projection of 128-dimensional embeddings clearly shows the progression from random initialization to organized 2D grid-like structure, while PCA (Appendix~\ref{AsecB}) points to the underlying curved manifold geometry. This self-organization demonstrates that geometric inductive biases emerge naturally from constraint satisfaction training.}
    \label{fig:grid_evo}
\end{figure*}

When visualizing low-dimensional projection of the static embeddings corresponding to individual points, we found that they organize themselves into a 2D grid they represent. In Figure \ref{fig:grid_evo}, we show how this organization emerges during training and how it is connected to the precision of the prediction (three dimensional projection is depicted in Appendix~\ref{AsecB}).

We stress that the existence of grid structure of the data domain and the existence of geometric figures related to constraints is given to a model only indirectly through the constraints and their solutions. The whole training dataset can be thought of as a set of constraints written in an unknown language, and the goal of training is to find a model for this language which will enable correct prediction. As the prediction is based on the similarity (given by inner product) of the dynamic embeddings of variables and static embeddings of points, it is not that unexpected that the discovered model reflects the geometry behind this language.

Here we focus on point embeddings, for completeness we provide some results about constraint embeddings in Appendix \ref{app:constraint_embeddings}. We show that constraint embeddings encode constraint types, geometric properties, satisfaction status, and temporal information up to some extent.

\subsection{Network Initialization with Grid Structure}
\label{sec:gridinit}

We investigated whether providing the model with an initial understanding of grid structure accelerates training convergence. Rather than initializing the shared embedding and classifier weights randomly, we can initialize them to reflect the geometric structure of the $20 \times 20$ grid.

To provide the model with an initial geometric inductive bias, we initialize the embedding matrix $\mathbf{W} \in \mathbb{R}^{N \times d}$ (used for both known point embeddings and the final classification layer) using a rotated grid construction. We begin by assigning each of the $N$ grid positions ($N=400$ for a $20 \times 20$ grid) its exact 2D coordinates $(x_i, y_i)$ and construct a matrix $\mathbf{V} \in \mathbb{R}^{N \times d}$ where $\mathbf{V}_{i,0} = x_i$, $\mathbf{V}_{i,1} = y_i$, and all other dimensions are zero. We then generate a random orthogonal matrix $\mathbf{Q} \in \mathbb{R}^{d \times d}$ via QR decomposition of a matrix with entries sampled from a standard normal distribution, and define initial embedding matrix as:
\[
\mathbf{W} = \mathbf{V} \mathbf{Q}
\]
This rotation spreads the spatial coordinate information across all $d$ dimensions while avoiding alignment with any particular coordinate axis. Following this initialization, training proceeds identically to the random case, with the weights in $\mathbf{W}$ remaining fully trainable.

\begin{figure}[ht]
    \centering
    \includegraphics[width=0.99\textwidth]{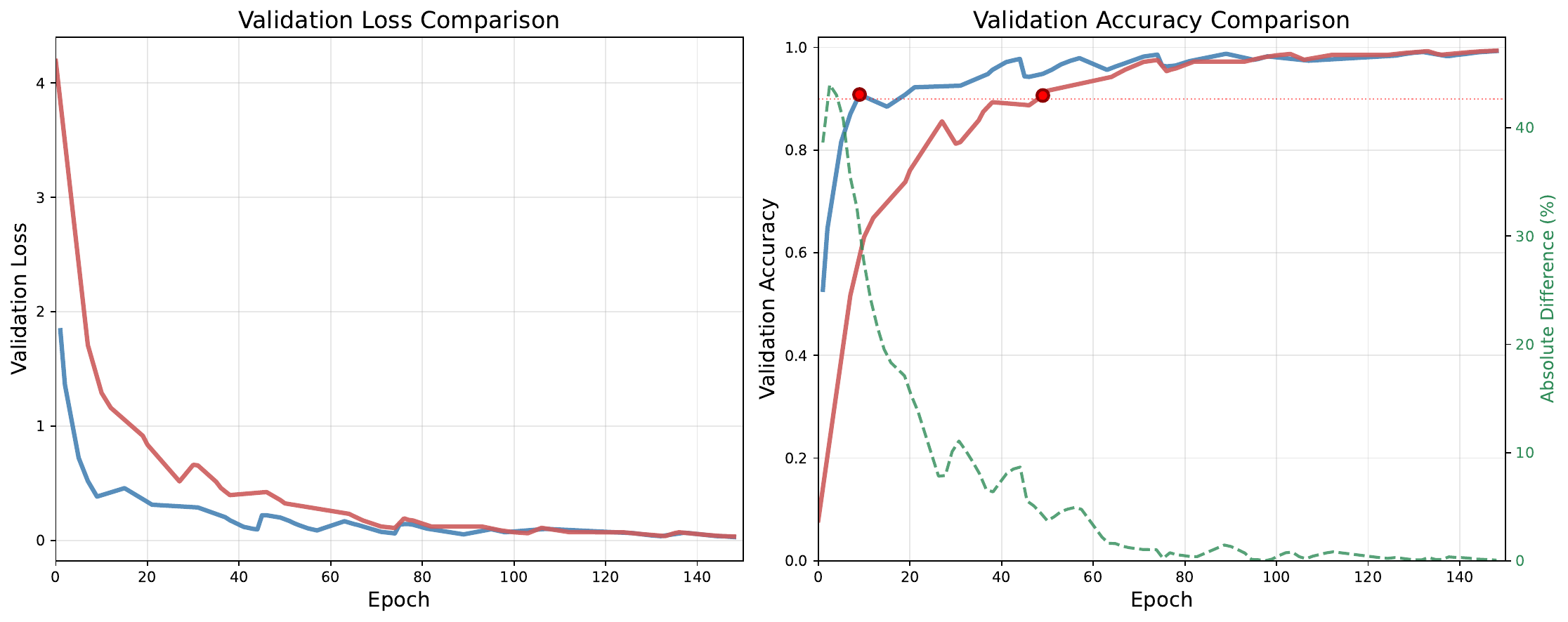}
    \caption{Comparison between random initialization of shared embedding and classification layer weights (red) and grid-structured initialization (blue) in terms of validation loss (left) and point accuracy (right). Red dots mark when each model first achieves 90~\% accuracy. Grid initialization reaches 90~\% accuracy within 10 epochs while random initialization requires almost 50 epochs. After 100 epochs their performance equalizes.}
    \label{fig:line22}
\end{figure}

Figure~\ref{fig:line22} demonstrates that grid-structured initialization provides substantial training acceleration compared to random initialization. The model with grid initialization converges to high validation accuracy within the first few epochs, while random initialization requires significantly more training time. This suggests that providing geometric inductive bias through weight initialization helps the model quickly discover the spatial relationships it needs to solve geometric constraint problems.

The initialization method shares weights between the embedding layer (used for known points) and classifier head (used for predictions), ensuring consistent geometric representations throughout the network. This architectural choice proves beneficial as the model can leverage the geometric structure for both encoding known positions and predicting unknown ones.

\subsection{Solution Process}\label{subs_sol_process}

In order to find the solution, the GNN model iteratively moves embeddings of unknown points in a high-dimensional space. During the forward pass, the same update rule is applied over and over again. Hence, it is possible to extract the embeddings in each message passing iteration to get a better understanding of the underlying process. More similar examples are provided in Appendix \ref{sec:appsolutionprocess}.

Figure \ref{fig:point_evo} shows an example of the solution process for a random problem given to the GNN. After each update, the closest static point embedding is taken for each variable of a problem and is visualized as the corresponding grid position.

 Note that in this example, the GNN first finds a close approximation to the hidden configuration and then refines it. Here, the squares are first ``approximated'' by quadrilaterals which then converge to exact squares. It can also be observed that one square constraint is approximately satisfied earlier than the other which reflects the fact that the other constraint can be resolved only after the first constraint is resolved (as explained in the next section). For an extended discussion and visualization of the reasoning dynamics as revealed by UMAP, see Appendix \ref{sec:umap-evolution}.
 
\begin{figure*}[ht]
    \centering
    \includegraphics[width=0.99\textwidth]{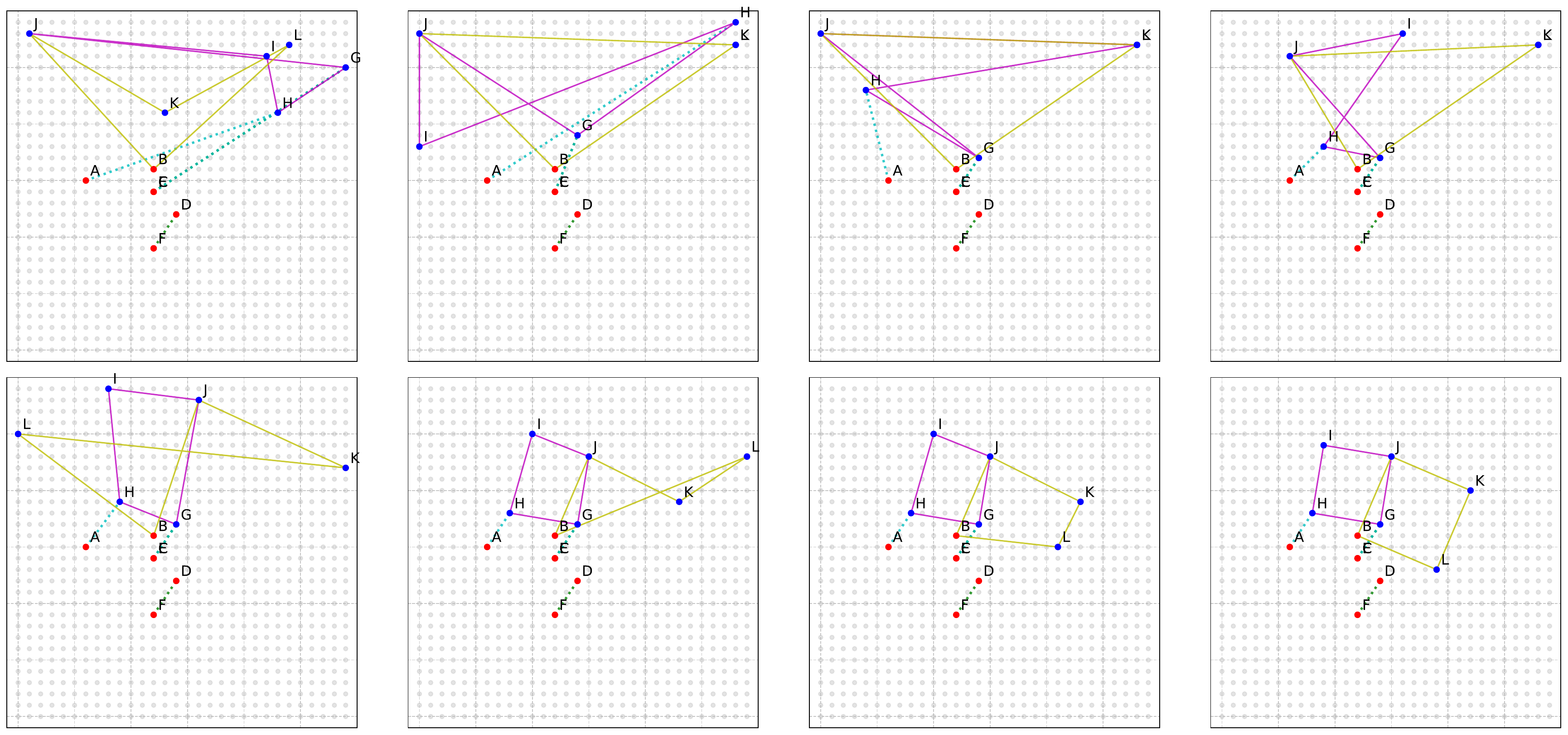}
    \caption{Visualization of the solution process. The red points A, B, C, D, E, F (C = E) are known, and the blue points G, H, I, J, K, L need to be predicted. There are two translation constraints: T (F, D, C=E, G), T (C=E, G, A, H) and two square constraints: S (H, I, G, J), S (B, J, L, K). Translations are marked by a dotted line and squares by a solid line. The network is trained to predict the result in $15$ iterations, of which initial state and results after iterations $3, 5, 7, 9, 11, 12, 13$ were chosen for illustration. The network gradually improves the result over the iterations: the first translation with only one unknown point G is solved, followed by finding the point H of the second translation. After translations, both squares are solved. Note this visualization was created with model operating on 30x30 grid.}
    \label{fig:point_evo}
\end{figure*}

\subsection{Analysis of Incorrectly Classified Points}\label{subs_err_analysis}

\begin{figure*}[ht]
  \centering
    \subfloat{\includegraphics[width=0.41\textwidth]{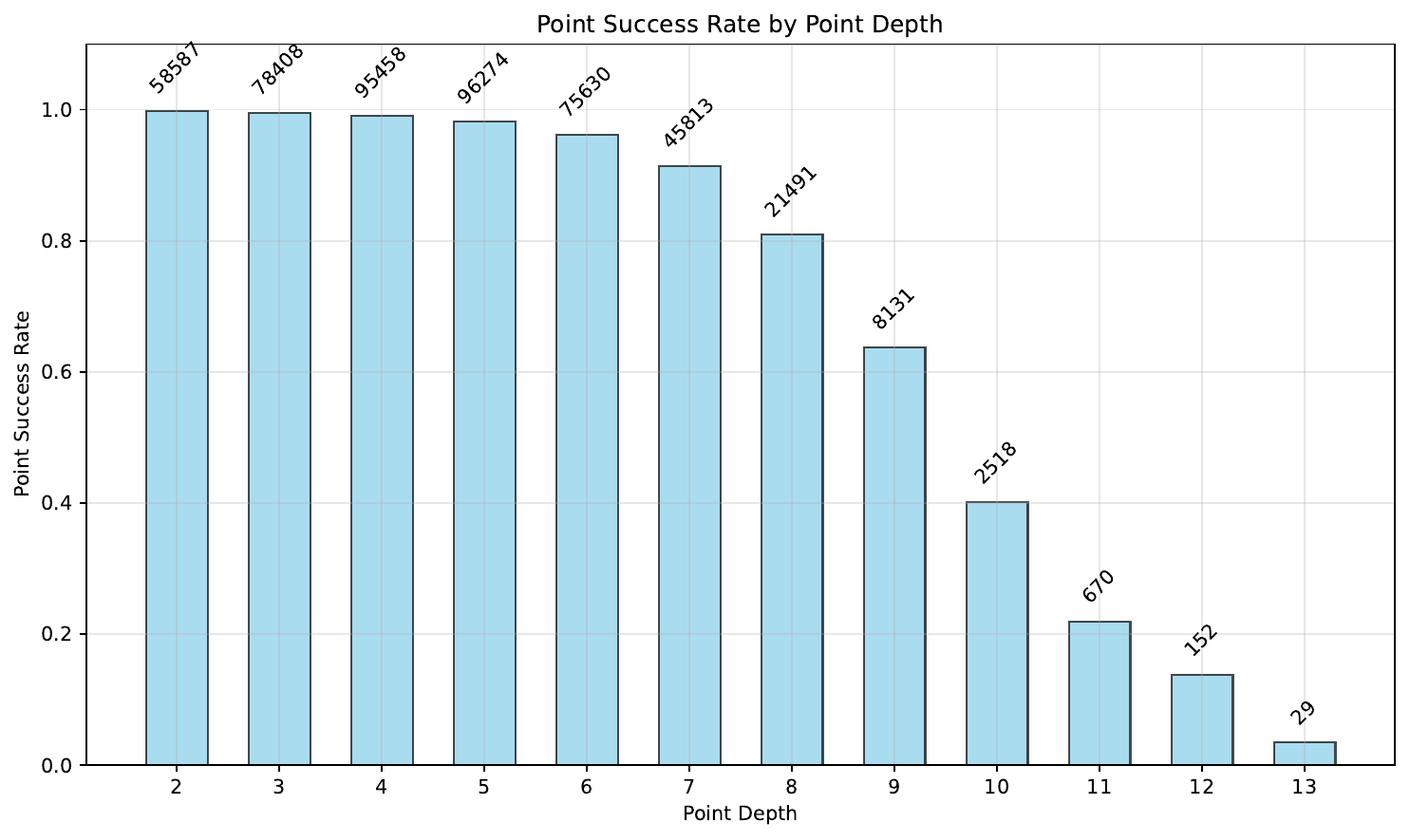}}
    \subfloat{\includegraphics[width=0.425\textwidth]{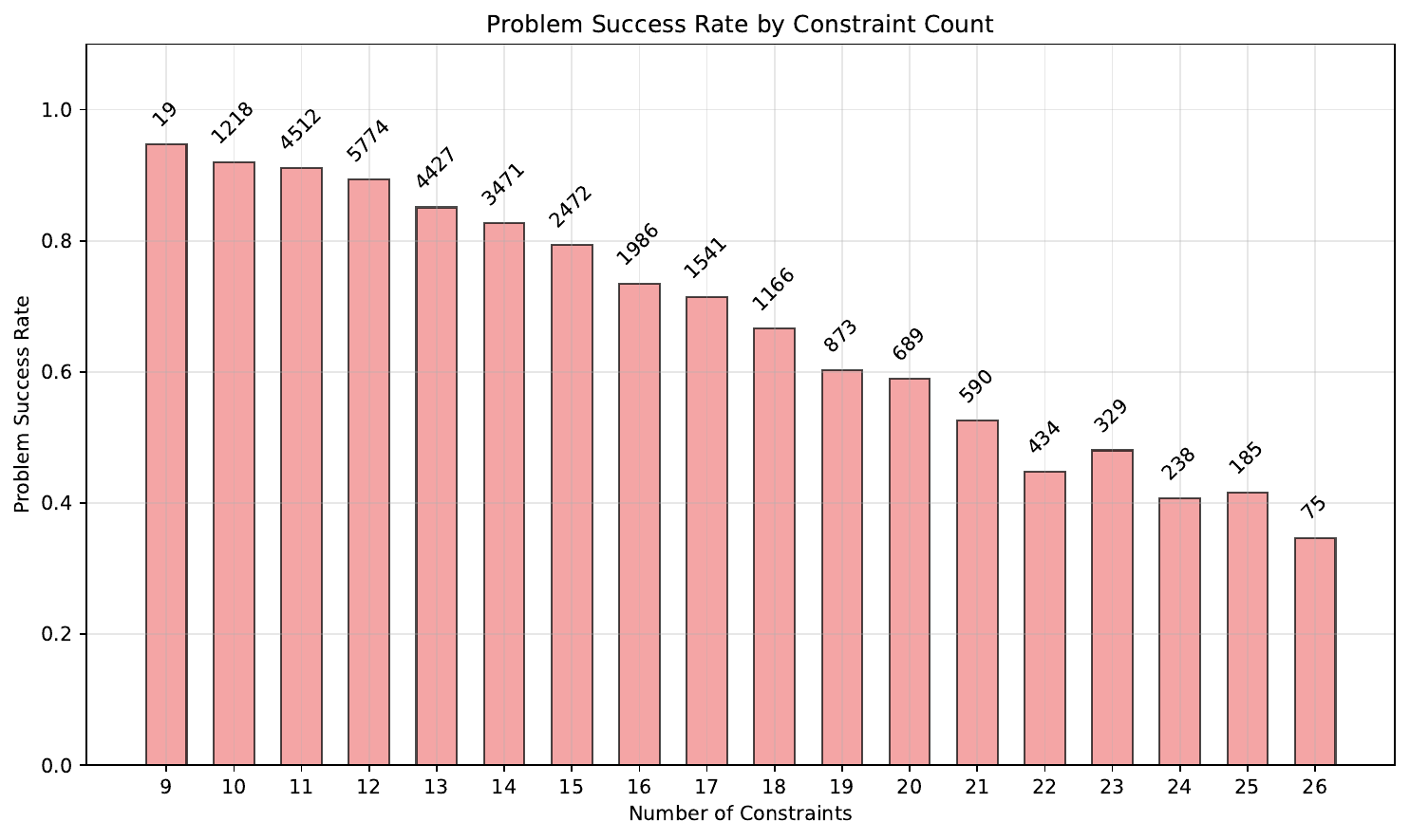}}
  \caption{Failure analysis on test dataset (15 iterations, 1 resample). Left: individual point prediction accuracy degrades with reasoning depth, showing points requiring deeper dependency chains have higher failure rates. Right: complete problem success decreases with constraint count. Numbers above bars show absolute sample sizes for each category.}\label{fig:depthrate}
\end{figure*}

To understand the failure modes of the GNN, we analyze predictions on our $30,000$ test instances using the basic model setup ($15$ iterations, $1$ resample) from Table~\ref{tab:spmd_results}. We examine how prediction accuracy relates to problem complexity along multiple dimensions.

As explained in Section~\ref{sec:data_gen}, constraints form a dependency structure where some must be resolved before others. For any given point, we can trace the dependency chain of constraints that must be resolved to determine its position. We define \emph{point depth} as the number of constraint resolution steps required for a specific point. Problems themselves vary in complexity through both the total number of constraints and the maximum depth of dependency chains within the problem.

Figure~\ref{fig:depthrate} (left) shows that prediction accuracy correlates negatively with point depth. Points requiring deeper reasoning chains have significantly higher failure rates. The relationship is nearly monotonic, with success rates declining from over $90\,\%$ for points at depth $2$--$3$ to below $20\,~\%$ for points requiring $10$ or more resolution steps. This pattern aligns with the iterative solution process observed in Section 
\ref{subs_sol_process} where the model progressively constructs solutions by first resolving simpler constraints.

Figure~\ref{fig:depthrate} (right) demonstrates that problems with more constraints also show lower complete success rates. This reflects both the increased likelihood of errors accumulating across multiple predictions and the tendency for larger problems to contain deeper dependency structures.

These findings suggest that the GNN's reasoning capability is fundamentally limited by the length of dependency chains it can reliably process. Importantly, when predictions do fail, the incorrectly classified points tend to be positioned close to their correct target locations, as evidenced by the distance histograms in Appendix~\ref{Adistanceerror}.

%% file: sections/limitations.tex
We note that we study a much simpler setup than was studied in AlphaGeometry. Predicting positions of points satisfying a set of geometric constraints is straightforward in comparison with predicting useful auxiliary points. Also, in our case, the unknown points need to lie on a discrete 2D grid so that we can train the model with a classification loss. The simplified setup turned out to be sufficient to study how NNs can reason about geometric constraints. Training the model for predicting auxiliary points requires a large computational budget and, moreover, the authors did not release the training dataset, which is also expensive to create.  We also mention that we use only four types of constraints, but the whole setup could be easily extended to different sets of constraints by adding more update functions in Equation~\eqref{eq:Z_update}. These simplifications enabled us to provide partial understanding of the process by which the tested models are able to solve individual problems and to demonstrate the benefits of using a GNN instead of a Transformer. We note that variable permutation invariance is not considered in our constraint definitions, as a fixed variable ordering is required to ensure unique solutions.

Compared to the GNN which Hula et al. \cite{hula2024revisiting} used for Boolean satisfiability, our GNN deals with fewer constraints and points. In their case, the GNN was able to handle hundreds of constraints, whereas our GNN deals with an order of magnitude fewer constrains and still needs more training samples than in the case of Boolean satisfiability. This could be caused by the fact that the domain of our constraints is much larger, but it could also be the case that different architecture of the GNN (with different update functions) could scale better to larger problems.

It is also possible that a different training procedure might result in significantly easier training. 
The inference process depicted in Figure \ref{fig:point_evo} and the findings of \cite{hula2024revisiting} suggest that the GNN could implicitly optimize an unknown objective during the iterative forward pass. Finding an expression for this objective could be interesting and used to train the network in an unsupervised way. This should be much easier than training it with the classification loss. Another obvious direction is to train the model as a diffusion model which ``denoises'' randomly assigned points to points of the solution.

Lastly, we mention that our experimental setup can be extended to more complex CSPs which could contain temporal relations and relations between various entities. We believe that such CSPs could be very useful for studying how NNs can generalize to unseen situations \cite{abbe2023generalization} and how they can discover models of the world without grounding \cite{Wong2023FromWM}. Studying geometric CSPs has the advantage that the domain has an obvious geometric interpretation which is visible also in the embedding space.

%% file: sections/fuwork.tex
\section{Future Work}
While we provided initial insights into geometric reasoning in neural networks, several directions remain for deeper understanding. Future work should investigate the expressiveness requirements of different update mechanisms, building on our finding that LSTMs significantly outperform RNNs for constraint updates. This includes determining the minimal architectural complexity needed to solve geometric constraints and whether these mechanisms can recover explicit mathematical formulas underlying constraints rather than learning approximations. The iterative refinement process that we observed suggests connections to continuous optimization methods. Future work should formalize this relationship and investigate whether training procedures based on explicit optimization objectives could improve efficiency. Additionally, extending beyond our four constraint types while addressing variable permutation invariance would broaden insights. Moving from discrete grid classification to continuous coordinate regression would enable analysis of geometric reasoning in unrestricted spatial domains.

%% file: sections/conclusion.tex
We have shown that GNNs as well as Transformers can learn to solve geometric CSPs and provided several insights into the process by which they find the solution. The visualizations show that when processing the problem, models form the hidden spatial configurations described by the problem in the embedding space. During training, the static embeddings of individual points in the 2D grid organize themselves within a 2D subspace and reflect the neighborhood structure of the grid. We showed that the occurrence of errors depends on problem complexity, like the number of steps required to resolve all variables. The models solve constraints through an iterative refinement process and exhibit test-time scaling capabilities, where allocating additional computational resources during inference improves performance on harder (out of training distribution) problem instances. Lastly, we showed that GNNs are much easier to train and can be scaled to significantly larger grids than Transformers.

%% file: sections/appendix.tex
\section{Training Procedures and Hyperparameters}\label{APhyper}

This section details the training procedures and hyperparameter configurations used for the GNN models, which were the primary focus of experiments.

\subsection{GNN Training Procedure}

The GNN uses AdamW optimizer with cosine annealing learning rate scheduling. The scheduler operates in 15-epoch cycles where each cycle's peak learning rate decreases by factor 0.9 from the previous cycle. Learning rate restarts cause temporary 0.5--6~\% accuracy (larger in initial stages of training) drops but enable continued optimization, with models regaining and exceeding previous performance within few epochs.

Exponential Moving Average (EMA) weights with decay 0.99 provide stable evaluation metrics. EMA weights are used exclusively for validation and testing while training continues with primary parameters. Cross-entropy loss applies only to unknown points, with known points remaining fixed.

The model supports the training with both fixed and changing number of iterations (number of message passing repetitions is different for each batch). Changing iteration training samples from a diminishing probability distribution centered at 15 iterations ($\pm$10 range), where 15 iterations occur 40~\% of the time, 16--17 iterations occur $\sim$15~\% combined, while extreme values like 25 iterations occur $<$1~\%. This technique improves test-time scaling robustness. Iteration count does not affect trainable parameters since the same layers are applied recurrently.

Dropout (0.1) applies to both point and constraint embeddings during training.

\subsection{Hyperparameter Selection for GNN}

Hyperparameter search determined that 15 training iterations outperform higher values like 20 for convergence speed and final accuracy. Smaller batch sizes (32) consistently outperform larger ones, trading GPU utilization for model accuracy. With random initialization, larger batch sizes prevent 99~\%+ convergence, plateauing around 90~\% after 200 epochs.

Grid-structured initialization enables broader hyperparameter ranges while maintaining performance, though the final configuration uses conservative settings reliable across initialization methods.

All hyperparameters were optimized for geometric constraint problems on 20$\times$20 grids. A summary for GNN as well as Transformer architecture used in initial comparison reported in Section \ref{subsComparison} are visible in Table \ref{tab:hyperparams}.

\begin{table}[ht]
\centering
\caption{Model hyperparameters and training settings for both architectures.}
\label{tab:hyperparams}
\begin{tabular}{lll}
\toprule
\textbf{Parameter} & \textbf{GNN} & \textbf{Transformer} \\
\midrule
\multicolumn{3}{l}{\textit{Model Architecture}} \\
Embedding dimension & 128 & 256 \\
Model iterations/layers & 15 $\pm$10 (diminishing) & 6 \\
Number of heads & -- & 6 \\
Dropout rate & 0.1 & -- \\
Positional embeddings & -- & RoPE \\
\midrule
\multicolumn{3}{l}{\textit{Training Configuration}} \\
Optimizer & AdamW & AdamW \\
Learning rate & $10^{-3}$ & $5 \times 10^{-4}$ \\
Weight decay & $10^{-3}$ & -- \\
Batch size & 32 & 512 \\
Epochs & 200 & 200 \\
Warmup steps & -- & 200 \\
\midrule
\multicolumn{3}{l}{\textit{Learning Rate Schedule}} \\
Scheduler & Cosine annealing & Linear \\
Cycle length & 15 epochs & -- \\
Peak decay factor & 0.9 & -- \\
Min LR factor & 0.1 & -- \\
\midrule
\multicolumn{3}{l}{\textit{Regularization}} \\
EMA decay & 0.99 & -- \\
Gradient clipping & 0.65 & 1.0 \\
\midrule
\multicolumn{3}{l}{\textit{Special Configuration}} \\
Special tokens & -- & [SEP], [UNK], [PAD], [MASK] \\
\bottomrule
\end{tabular}
\end{table}

\subsection{Model Complexity}

For 20$\times$20 grids with embedding dimension 128, the GNN contains 1,498,112 trainable parameters. Each constraint type contributes 328,704 parameters to the variable - constraint ($U_c$) message passing layers. The shared embedding and classifier layers account for 51,200 parameters (400 grid positions $\times$ 128 dimensions), counted once since they share the same weight matrix. The constraint - variable ($U_x$) message passing layer contributes 132,096 parameters. In comparison, the Transformer model for the same grid size contains 5,081,088 parameters, approximately 3.4 times larger than the GNN architecture.

\subsection{LSTM vs. RNN Constraint Update Ablation}\label{app:rnn_ablation}

We conducted an ablation experiment to evaluate whether our constraint update mechanism benefits from using an LSTM cell over a simpler RNN cell. While the main experiments use LSTM-based updates, we performed a basic hyperparameter search for the RNN variant, adjusting learning rate, embedding dimension, batch size, and number of iterations. Across all runs, the RNN remained unstable and failed to exceed 40\% validation accuracy. When using the same hyperparameters as the LSTM model (for a direct comparison), the RNN plateaued at 38.4\% and exhibited higher validation loss.

\begin{figure}[ht]
    \centering
    \includegraphics[width=0.9\textwidth]{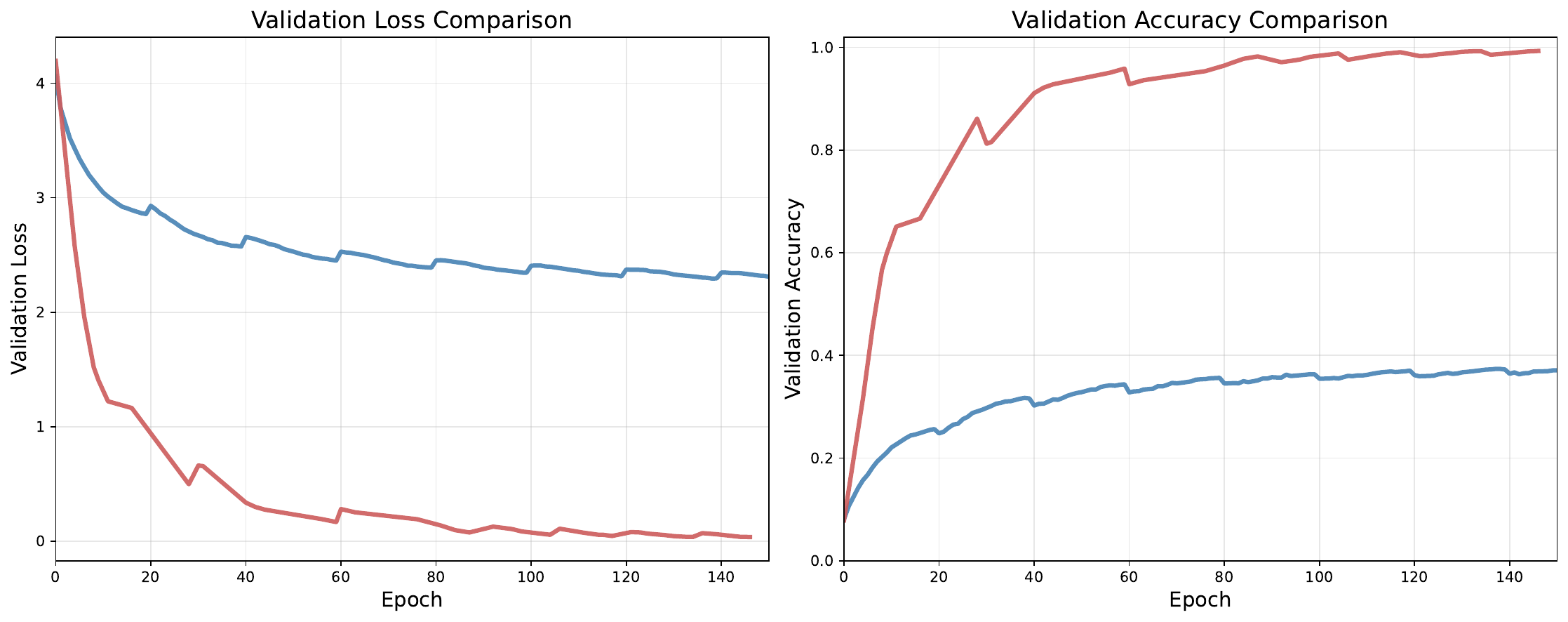}
    \caption{Comparison of LSTM and RNN-based constraint update mechanism. The LSTM achieves higher validation accuracy and lower loss (red). The RNN variant achieves only 38.4\% accuracy and performs worse in terms of loss (blue).}
    \label{fig:train_rnn_comparison}
\end{figure}

Figure~\ref{fig:train_rnn_comparison} shows the validation accuracy and loss during training for both variants. These results confirm the advantage of using more expressive update mechanisms (such as LSTM) for modeling our geometric constraints.

\section{Embedding Visualization for the GNN in 3D}\label{AsecB}

As mentioned in Section~\ref{sec:vis}, the static embeddings of individual points self-organize into a grid structure representing their spatial relationships. This section provides 3D visualizations to complement the 2D projections shown in the main text.

\begin{figure}[ht]
    \centering
    \includegraphics[width=0.8\textwidth]{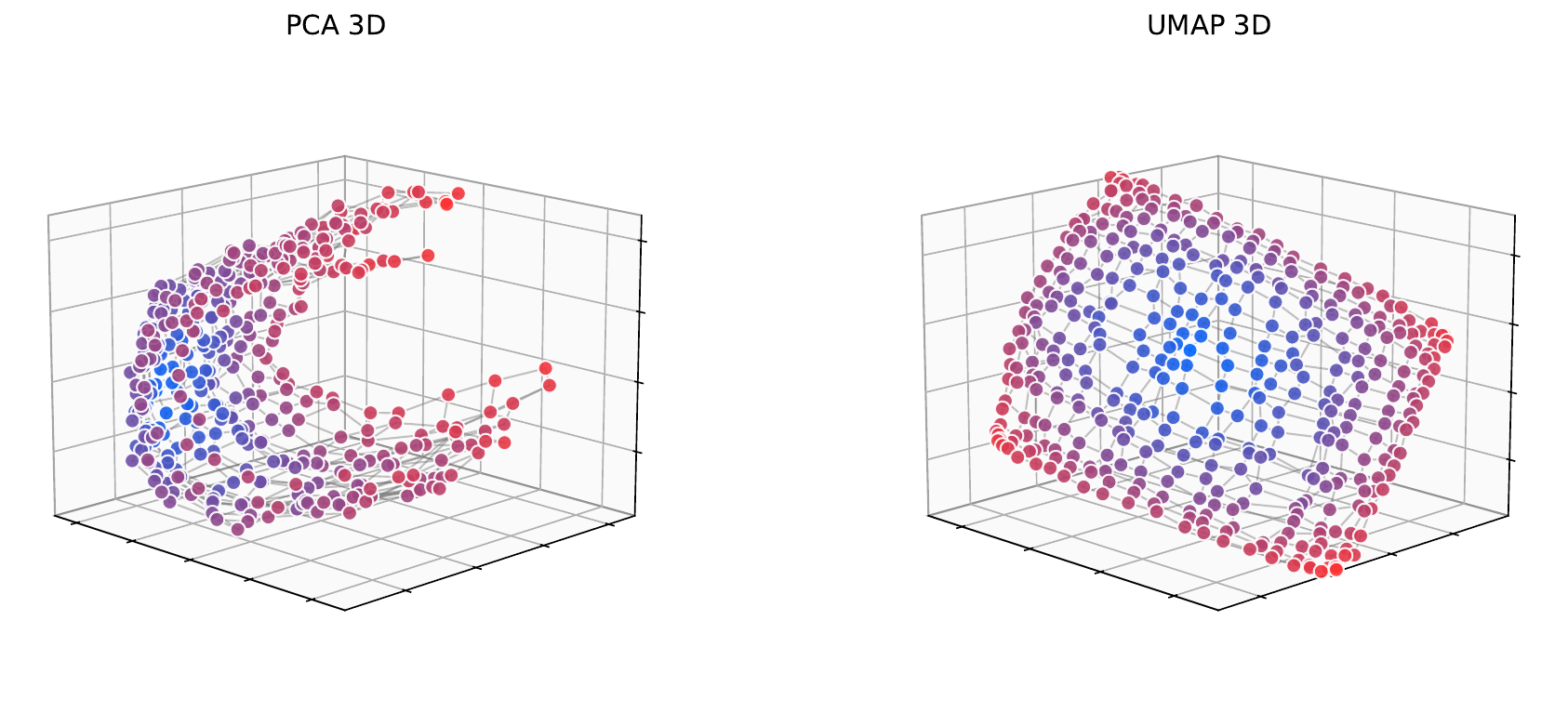}
    \caption{3D projections of static point embeddings from 128-dimensional space using PCA (left) and UMAP (right). PCA reveals a curved "cup" or "bell" shaped structure, while UMAP projects them onto a flatter surface that more clearly shows the 2D grid organization. Colors indicate distance from grid center.}
    \label{fig:tformer_gridG3D2}
\end{figure}

Figure~\ref{fig:tformer_gridG3D2} shows 3D projections of the learned embeddings using both UMAP and PCA methods. From UMAP projections we observed that during training, randomly initialized embeddings evolve from a single spherical cluster, which later unfolds into a U-shaped surface and finally converges to a flat 2D surface.

In contrast, the PCA 3D visualization remains a curved "cup" or "bell" shape rather than the flat plane observed with UMAP even after full training. Thus the embeddings lie on a curved surface in the high-dimensional space rather than in a flat plane.

\begin{figure}[ht]
    \centering
    \includegraphics[width=0.8\textwidth]{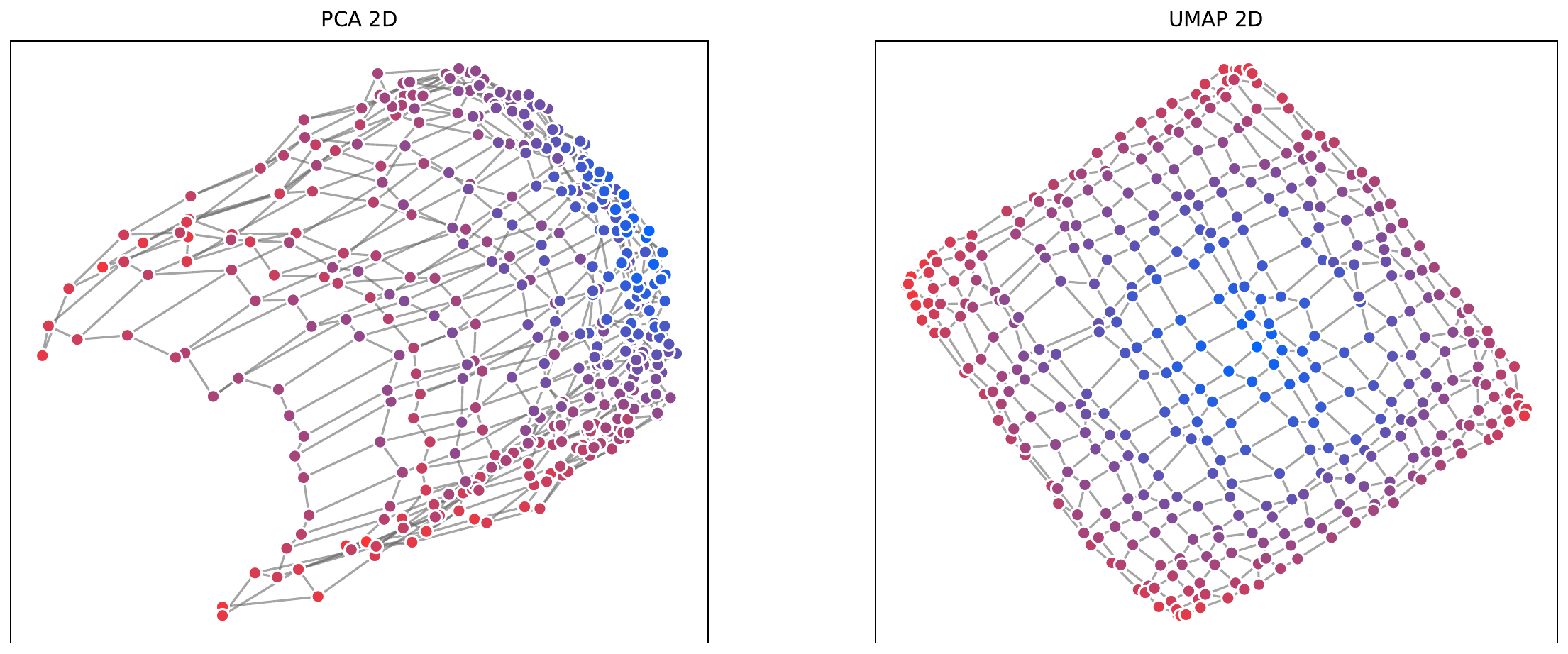}
    \caption{2D projections of static point embeddings from 128-dimensional space using PCA (left) and UMAP (right). PCA shows curved, warped structure while UMAP better preserves the regular grid connectivity. Lines show spatial adjacency relationships from the original 20$\times$20 grid. Colors indicate distance from grid center.}
    \label{fig:tformer_gridG2D}
\end{figure}

\begin{figure}[h!t]
    \centering
    \includegraphics[width=0.85\textwidth]{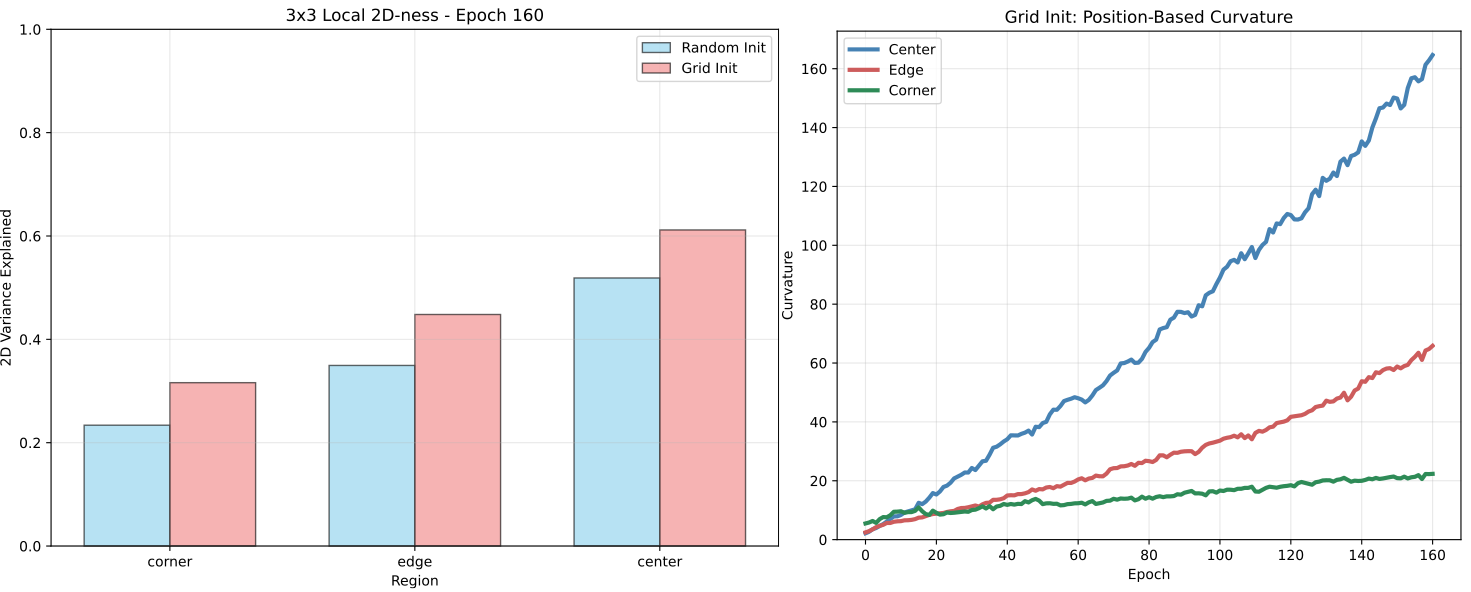}
    \caption{Curvature and local 2D-ness during training for random vs. grid initialization. Right: Mean curvature over training time and by spatial region. Curvature is computed in the 3D PCA projection using a local neighborhood eigenvalue method, and it reflects how much local patches deviate from planarity. The rising curvature over time corresponds to the emergent cup-like shape seen in 3D PCA plots. Left: 2D-ness, measured as the fraction of variance explained by the top-2 PCA components in $3\times3$ spatial subgrids.}
    \label{fig:curvature}
\end{figure}

To further characterize the geometric evolution of the embedding space, we measure local curvature and local dimensionality throughout training (\autoref{fig:curvature}). 

Curvature is computed in the 3D PCA projection using an eigenvalue-based method that quantifies deviation from local planarity within small neighborhoods. Specifically, we compute the local covariance matrix \( C \in \mathbb{R}^{3 \times 3} \) of each neighborhood and define curvature as the ratio between the smallest and largest eigenvalues:
\[
\kappa = \frac{\lambda_{\min}(C)}{\lambda_{\max}(C) + \varepsilon},
\]
where \( \lambda_{\min} \) and \( \lambda_{\max} \) are the smallest and largest eigenvalues of \( C \), and \( \varepsilon \) is a small constant added for numerical stability.

In the case of random initialization, curvature starts relatively high and fluctuates. In contrast, curvature under grid initialization starts near zero but increases steadily over time, eventually reaching levels similar to the random case. This indicates that even when training begins from a perfectly planar manifold, the resulting structure develops nontrivial curvature as training progresses. In parallel, we evaluate local 2D-ness, defined as the variance explained by the top two principal components in $3{\times}3$ spatial subgrids. This captures how locally planar the embedding remains. Together, these metrics provide complementary views on how spatial structure and geometric complexity evolve, and how these differ across initialization strategies.

\begin{figure}[h!t]
    \centering
    \includegraphics[width=0.85\textwidth]{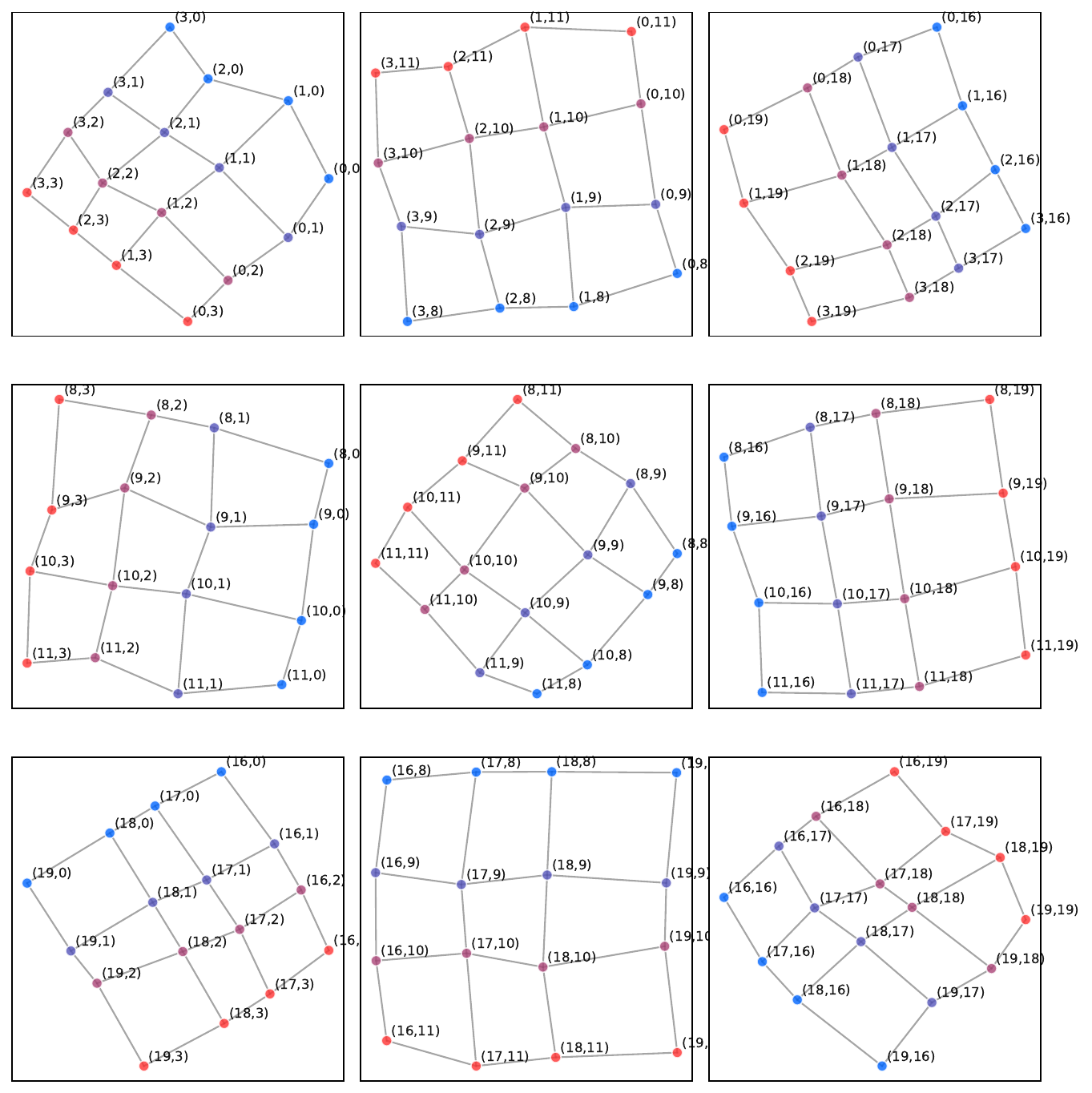}
    \caption{PCA projections of representative 4$\times$4 subregions from the 20$\times$20 grid embeddings, selected from corners and center areas for complete coverage. Local grid structure is well-preserved across all regions. While global embedding organization exhibits curvature (Figure~\ref{fig:tformer_gridG3D2}), these local neighborhoods maintain linear spatial relationships.}
    \label{fig:pca_subgrids}
\end{figure}

Figure~\ref{fig:pca_subgrids} demonstrates that when examining 4$\times$4 subregions of the 20$\times$20 grid, PCA projections reveal well-organized local grid structures. This visualization technique shows that the embeddings maintain locally linear relationships within smaller regions, even though the global structure exhibits curvature.

These visualizations confirm that the model successfully learns to embed spatial relationships in its high-dimensional representation, with the embeddings organized on a curved surface that preserves local neighborhood structure.

\begin{figure}[ht]
    \centering
    \includegraphics[width=0.9\textwidth]{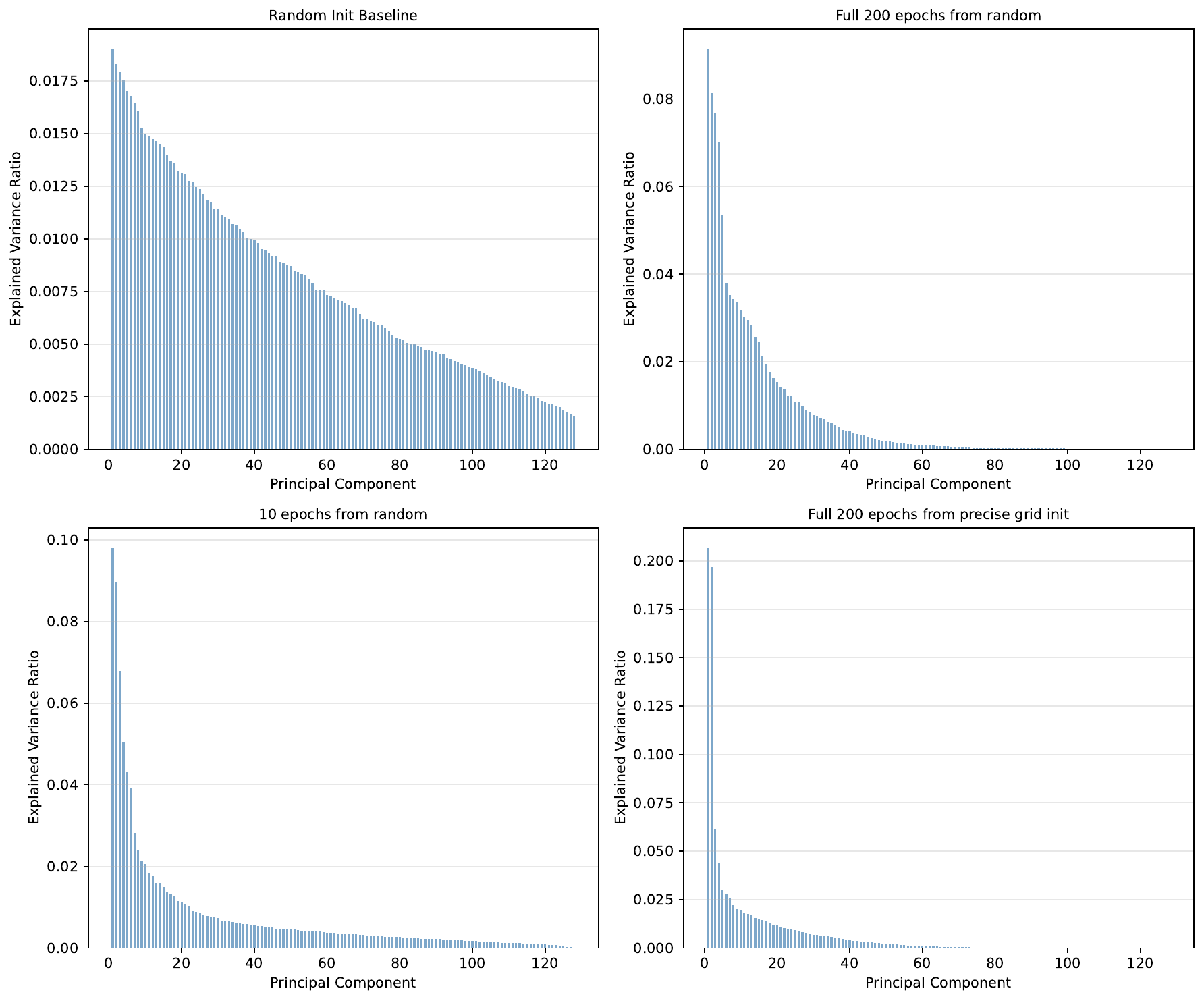}
    \caption{PCA component variance distribution across training conditions. Top left: random initialization baseline. Top right: after 200 epochs from random initialization, most variance spreads across approximately 5 components. Bottom left: early training (10 epochs from random) shows first 2 components are more prominent. Bottom right: after 200 epochs from precise grid initialization, variance utilizes more dimensions than baseline but less prominently than random initialization training.}
    \label{fig:pca_variance}
\end{figure}

Figure~\ref{fig:pca_variance} shows the distribution of variance across PCA components at different training stages and initialization methods. When training from random initialization, the first two components are most prominent at early stages (10 epochs). After full training (200 epochs), the variance spreads across approximately the first 5 components.

Training from precise grid initialization shows a different pattern. After 200 epochs, the first two components remain most prominent while the model utilizes additional dimensions, contrasting with the pure grid initialization baseline which concentrates variance primarily in two dimensions.

This analysis reveals how the usage of the embedding dimensions evolves during training and how initialization strategy affects the final embedding structure.

It also highlights the advantage of UMAP for our setup. UMAP offers a more accurate visualization of the learned grid structure than PCA. PCA is a global linear method and can distort local relationships when the embedding manifold becomes curved or non-planar, often placing distant points close together in projection. UMAP, by contrast, is a locally nonlinear method that better preserves neighborhood structure. This makes it more effective at representing the underlying grid organization when it becomes embedded in a non-planar, high-dimensional space.

\section{Distribution of the Error Magnitudes}\label{Adistanceerror}
\begin{figure}[ht]
  \centering
    \includegraphics[width=0.6\textwidth]{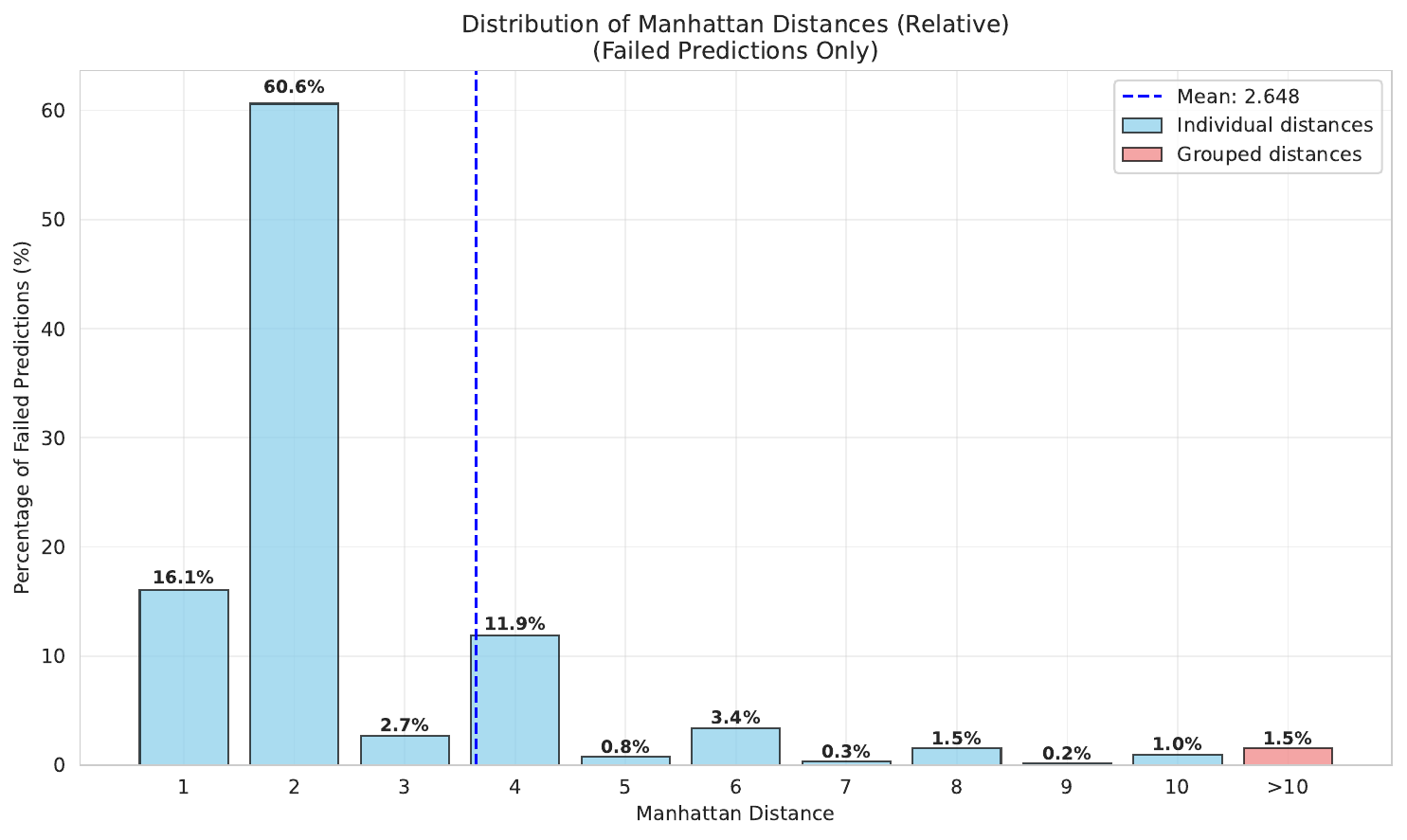}
  \caption{Distribution of prediction errors for incorrectly classified points, measured by Manhattan distance. Most failed predictions lie close to their correct positions.}\label{fig:euclidfail}
\end{figure}

The histogram in Figure~\ref{fig:euclidfail} depicts a distribution of Manhattan distances between the incorrectly predicted point position and the ground truth position on the grid. As can be seen, most incorrectly predicted positions lie very close to the ground truth position.

\section{Embedding Visualization for the Transformer Model}\label{AsecC}
\begin{figure}[ht]
    \centering
    \includegraphics[width=0.65\textwidth]{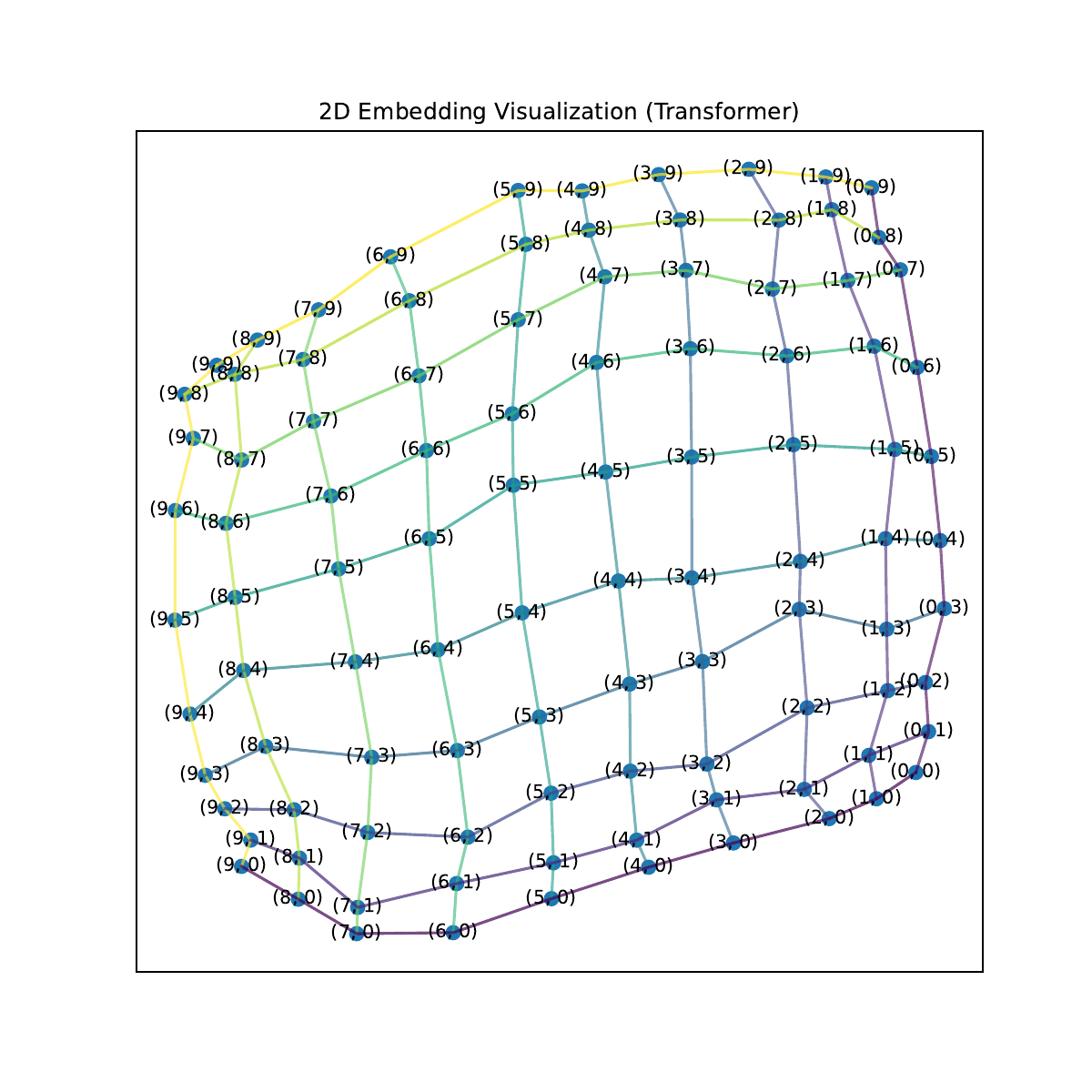}
    \caption{Projection into 2D using UMAP illustrates that static token embeddings of the Transformer self-organized into a grid-like structure which the visualized tokens represent. The edges show the connectivity of the points in the original grid.}
    \label{fig:tformer_grid}
 \end{figure}

In Figure~\ref{fig:tformer_grid}, we show the grid-like structure of static point embeddings discovered by the Transformer. In comparison with Figure~\ref{fig:grid_evo} 
which shows the emergence of the corresponding structure
for GNN, the grid is now of smaller size, namely 10$\times$10, since the Transformer was not able to scale into larger sizes with sufficient accuracy as discussed in Section \ref{subsComparison}.

\section{Chain-of-Thought Training}
\label{AsecCoT}

In order to train the Transformer to produce a chain-of-thought which assigns the variables incrementally, we implemented a simple solver which logs its steps and the resulting logs are used for imitation. The solver first orders the constraints according to a DAG mentioned in Section \ref{sec:data_gen} and then resolves the constraints one by one, starting from the root constraints. As the solver traverses the DAG, it logs the constraint of a given node, the values of already assigned variables within the constraint and lastly the computed values for the remaining variables. We also include few keywords into the log which delimit the provided information. Example of the log for a random problem is shown below:\\ 

\begin{minipage}[b]
{0.89\textwidth}
TRANSLATION ( 1 0 2 3 ) , TRANSLATION ( 5 4 7 6 ) , SQUARE ( 8 7 9 3 ) , SQUARE ( 11 8 10 3 ) , TRANSLATION ( 8 7 12 3 ) ;
fixed 0 = \#696 , 1 = \#617 , 2 = \#978 , 4 = \#577 , 5 = \#498 , 6 = \#731 ;
Solution begins ;
Con TRANSLATION ( 5 4 7 6 ) ;
Known 5 = \#498 , 4 = \#577 , 6 = \#731 ;
Impl 7 = \#652 ;
Con TRANSLATION ( 1 0 2 3 ) ;
Known 1 = \#617 , 0 = \#696 , 2 = \#978 ;
Impl 3 = \#1057 ;
Con SQUARE ( 8 7 9 3 ) ;
Known 7 = \#652 , 3 = \#1057 ;
Impl 8 = \#462 , 9 = \#867 ;
Con SQUARE ( 11 8 10 3 ) ;
Known 8 = \#462 , 3 = \#1057 ;
Impl 11 = \#247 , 10 = \#842 ;
Con TRANSLATION ( 8 7 12 3 ) ;
Known 8 = \#462 , 7 = \#652 , 3 = \#1057 ;
Impl 12 = \#867 ;
Solution ends\\{}
\end{minipage}\\Variables are expressed by a number (1, 2, 3, etc.) and individual points are expressed with point ID with a \# symbol in front (\#696, \#617, etc.). The input has two parts, a problem statement and a solution. We train on the whole input, but exclude the problem statement from the computation of the loss. For validation, we include only the problem statement. 
The keyword \emph{Con} marks the selected constraint, \emph{Known}
marks the known variables which appear in the selected constraint with their values and \emph{Impl} marks the newly deduced variables with their values.

\section{Test-Time Scaling and Iteration Analysis}\label{AsecTestTime}

This section analyzes model behavior across different iteration counts and resampling strategies, supporting the test-time scaling results presented in Table~\ref{tab:spmd_results}.

\begin{figure}[ht]
    \centering
    \includegraphics[width=0.99\textwidth]{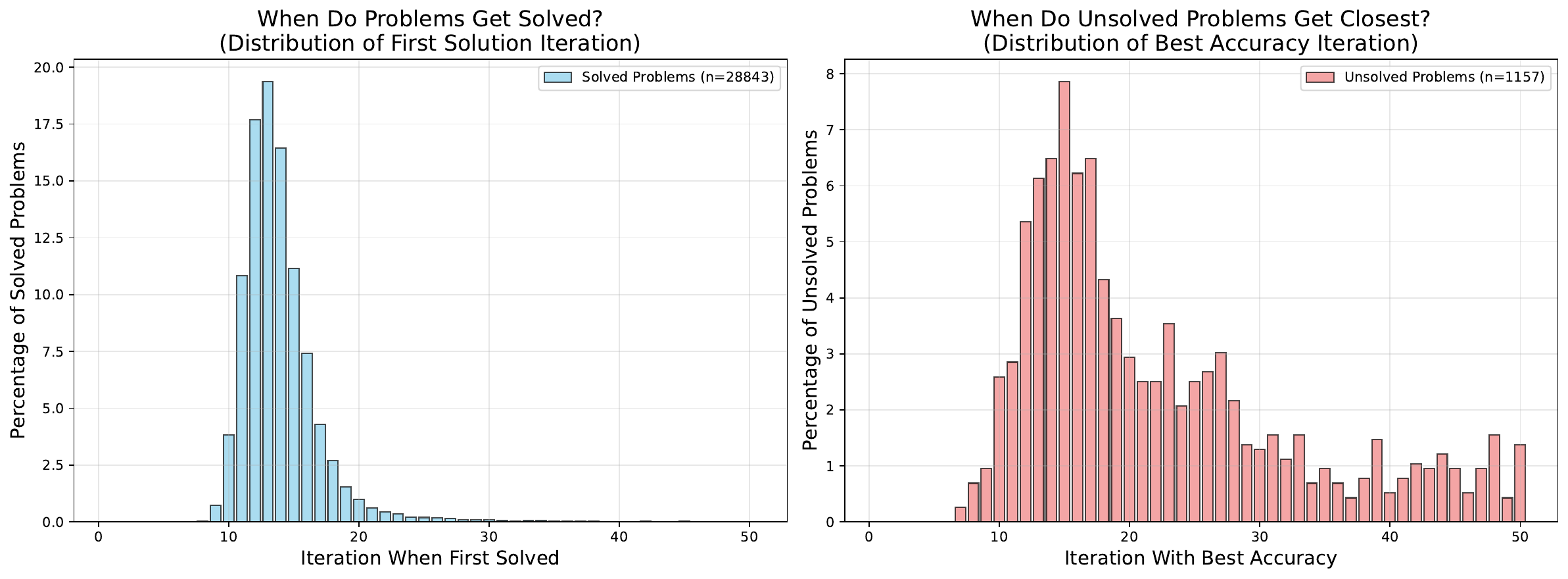}
    \caption{When problems reach solution during inference. Left: distribution of first solution iteration for successfully solved problems. Right: iteration when unsolved problems achieve their highest point accuracy.}
    \label{fig:whensolved}
\end{figure}

Figure~\ref{fig:whensolved} shows the distribution of iterations when variables are correctly assigned for the first time (left) and when unsolved problems achieve their best point accuracy (right) using single resampling. Most problems that can be solved are resolved within the first 15 iterations, with a long tail extending to 50 iterations. For unsolved problems, the peak occurs around iterations 10-15, though substantial numbers of problems achieve their best accuracy at later iterations, indicating that additional computation can still provide benefits.

\begin{figure}[ht]
    \centering
    \includegraphics[width=0.99\textwidth]{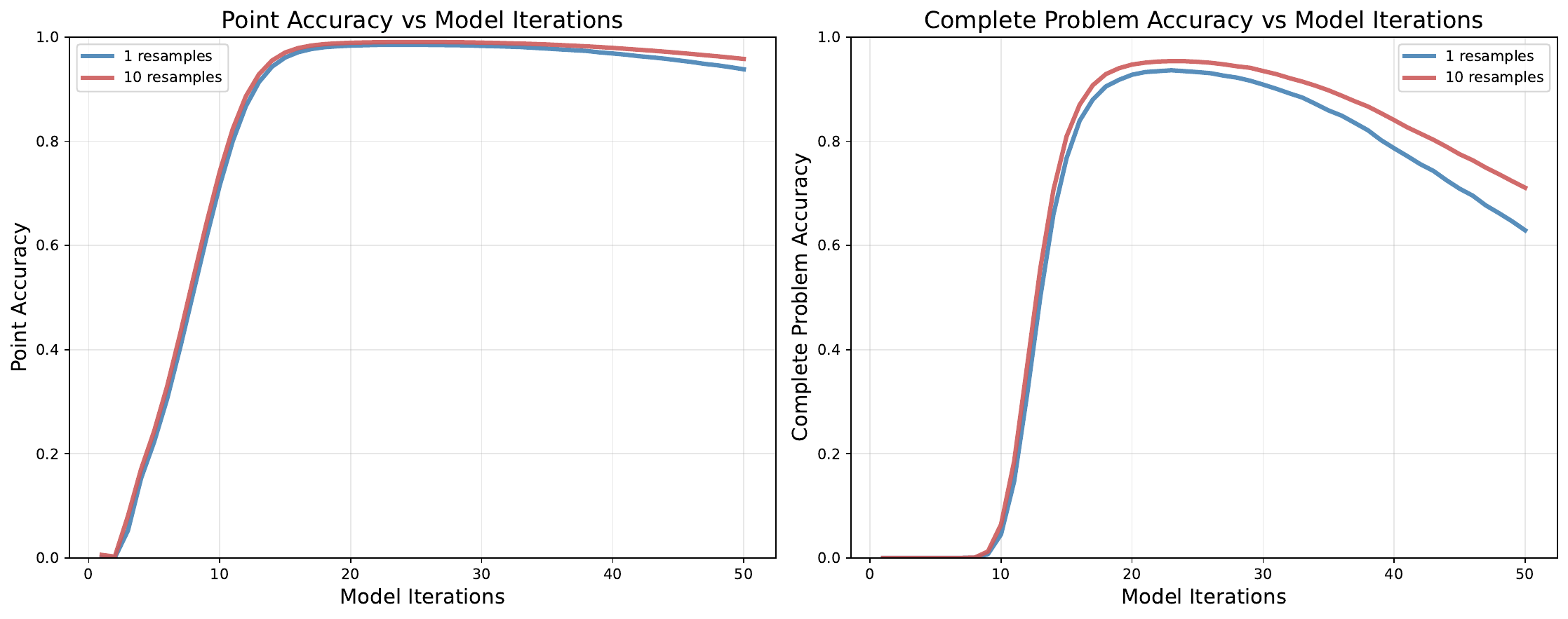}
    \caption{Model performance across iteration counts with single resampling (blue) and 10 resamples (red). Both point accuracy and complete problem accuracy peak around iterations 23-25, then decline. Multiple resampling provides consistent benefits across all iteration counts.}
    \label{fig:line1}
\end{figure}

Figure~\ref{fig:line1} demonstrates how accuracy evolves with iteration count for both single and multiple resampling strategies. Both point and complete problem accuracy peak around iterations 23-25, then decline with further iterations. This indicates that while some individual problems benefit from extended computation (as shown in Figure~\ref{fig:whensolved}), increasing iterations beyond 25 breaks more already-solved instances than it helps, resulting in net performance degradation.

Multiple resamples provide consistent benefits across all iteration counts, with optimal performance occurring in the 23-25 iteration range for both strategies.

These results explain why increasing iterations from 15 to 23 improves complete accuracy and why the "Best" oracle configuration achieves higher performance by selecting optimal iteration counts per problem.

\section{Scaling the Size of the Grid}\label{sec:scaling_laws}
To get a sense of how the sample complexity depends on the size of the grid, we train several models on different sizes of the grid and different amount of training samples. To achieve faster training, we conducted these experiments with problems which contained only two types of constraints ($S$ and $T$). 

The problem generator mentioned in Section 
\ref{sec:data_gen} 
produces problems which have on average around four constraints. Both types of constraints ($S$ and $T$) are sampled with equal probability. Therefore, we can assume that an average problem has two constraints of type $S$ (which is determined by $2$ points) and two constraints of type $T$ (which is determined by $3$ points). If we 
denote the number of points on the side of the grid by $n$, then we can estimate the number of unique problems in the grid of size $n \times n$ to be 
$n^{20}$. 
There are $n^2$ possible point positions and we independently sample 6 points for two constraints of type $T$ and 4 points for two constraints of type $S$, together yielding $(n^2)^{10}$ possibilities each of which determines one instance. This number should be viewed as an upper bound because it ignores the fact the constraints can share variables. %

To test the dependence of the validation accuracy on the grid size and number of training samples, we use the following values of $n$: $10$, $20$, $30$, $40$, $50$, $60$, $70$, $80$. This results in the following number\footnote{After rounding the exponent to the nearest integer.} of possible problems for each $n$, respectively: 
$10^{20}$, $10^{26}$, $10^{29}$, $10^{32}$, $10^{34}$, $10^{35}$, $10^{37}$, $10^{38}$. 

\begin{figure*}[ht]
  \centering
    \subfloat{\includegraphics[width=0.46\textwidth]{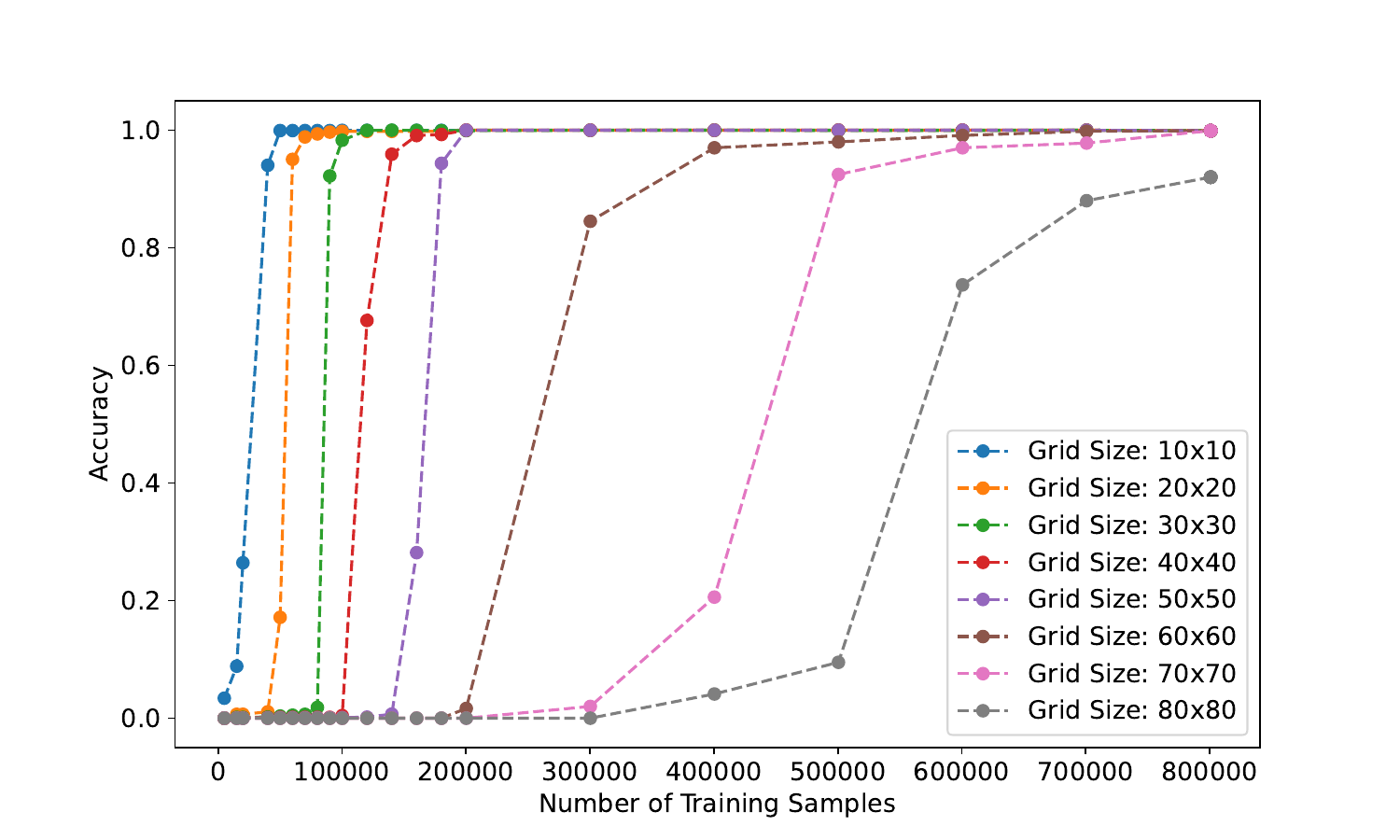}}
    \subfloat{\includegraphics[width=0.46\textwidth]{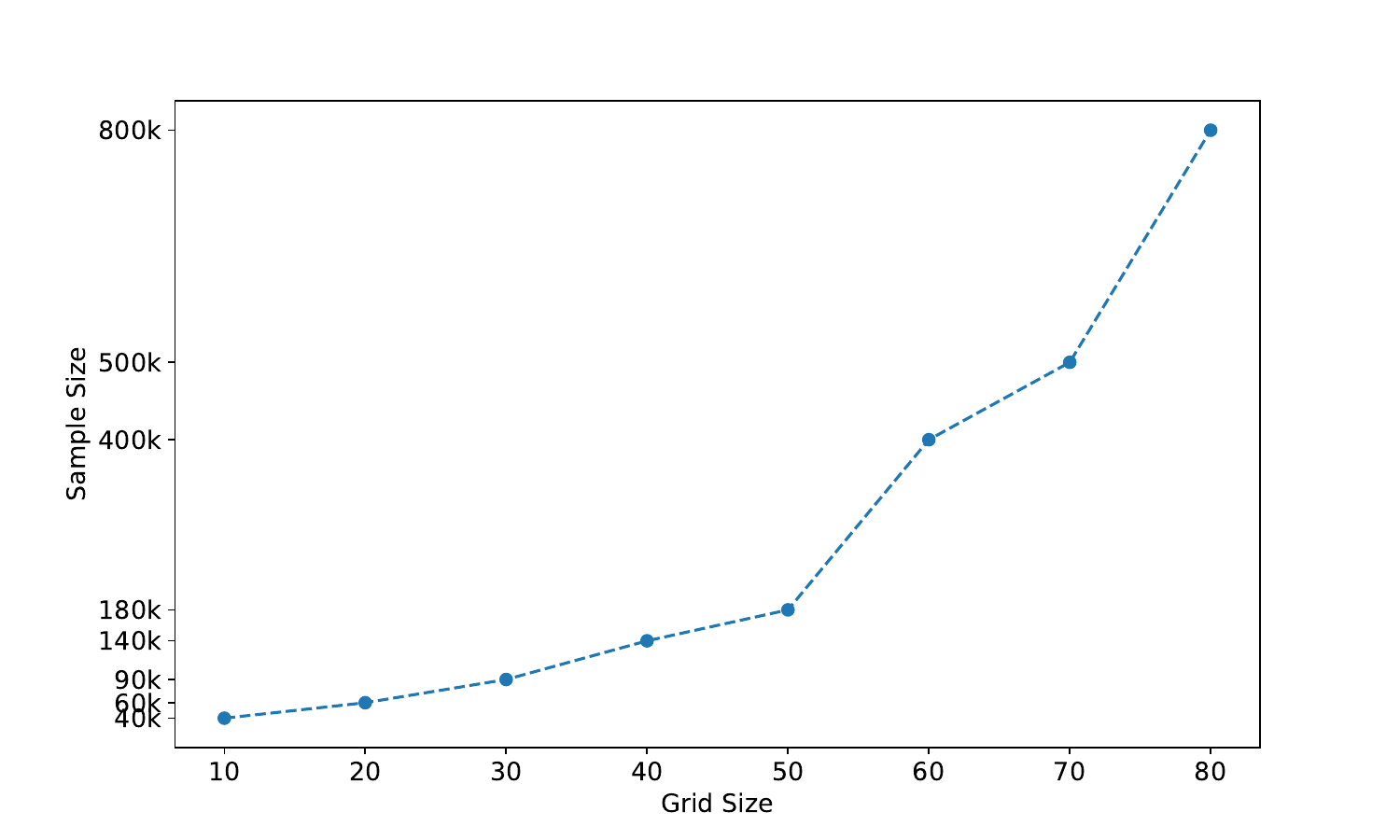}}
  \caption{Scaling analysis for different grid sizes using simplified problems with only Square and Translation constraints. Left: validation accuracy versus training set size for grids from 10$\times$10 to 80$\times$80 points. Right: sample complexity required to achieve 90~\% accuracy across different grid sizes.}\label{fig:scaling_plot}
\end{figure*}

The sizes of the training set for each grid size are in the range from $5$k to $800$k. The relationship between the validation accuracy, the grid size and the size of the training set can be seen in Figure \ref{fig:scaling_plot}. 
In the same Figure (right), we plot the relationship between the grid size and sizes of the training set for which the validation accuracy exceeded $90\,~\%$.

\section{More Examples of the Solution Process}\label{sec:appsolutionprocess}

This section provides additional examples of the iterative solution process to support the findings presented in Section~\ref{subs_sol_process}. While the main text focused on a single detailed example, these three instances demonstrate that the observed iterative refinement behavior is consistent across different problem configurations and constraint combinations.

\begin{figure}[h!t]
    \centering
    \includegraphics[width=0.85\textwidth]{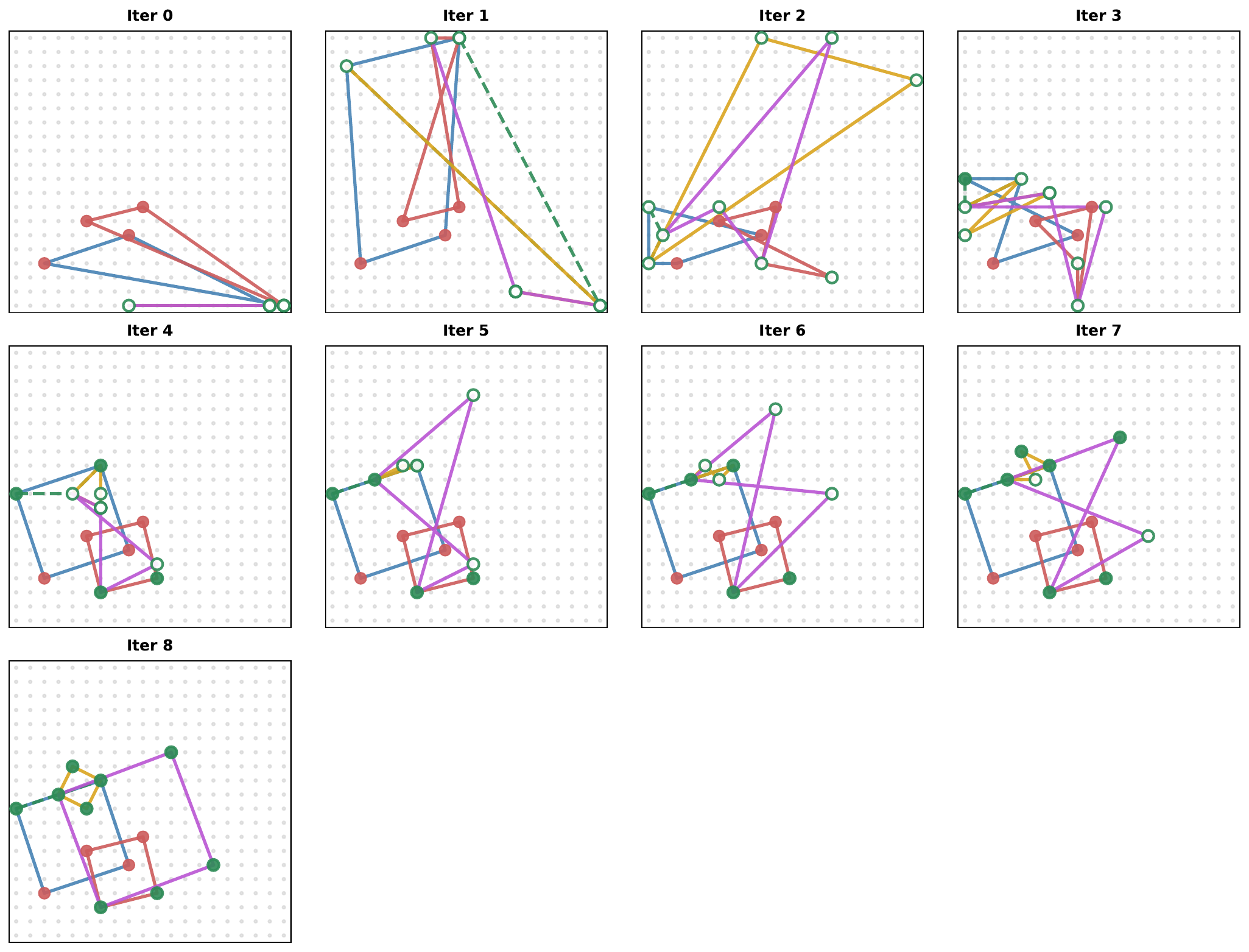}
    \caption{Problem with 5 constraints: four squares and one midpoint. The midpoint constraint (dashed green line) coincides with the top side of the blue square (two vertices of smallest square are its part).}
    \label{fig:sol1}
\end{figure}

\begin{figure}[h!t]
    \centering
    \includegraphics[width=0.85\textwidth]{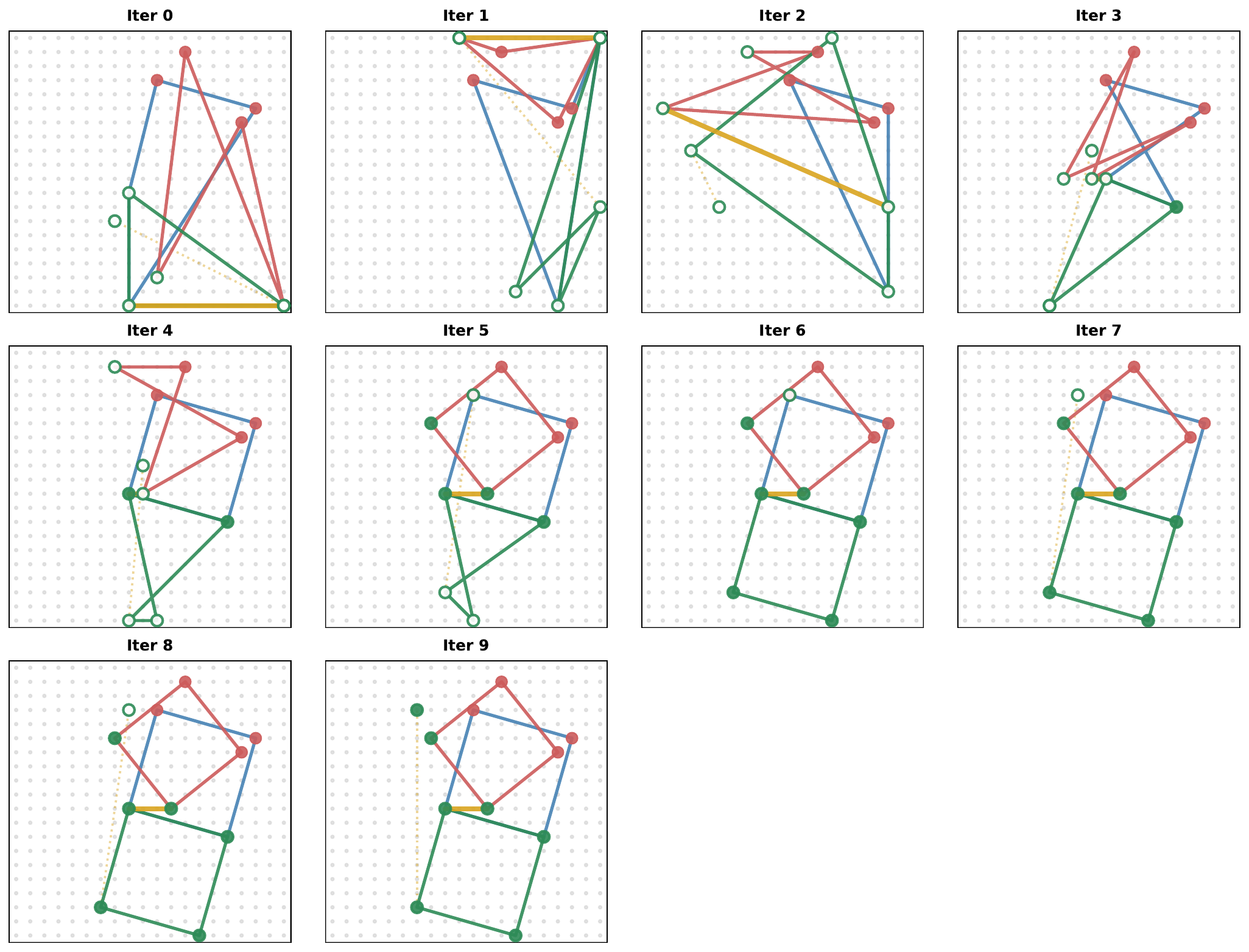}
    \caption{Problem with 4 constraints: three squares and one reflection. The two leftmost points are reflected across the axis formed by the yellow line segment.}
    \label{fig:sol2}
\end{figure}

\begin{figure}[h!t]
    \centering
    \includegraphics[width=0.85\textwidth]{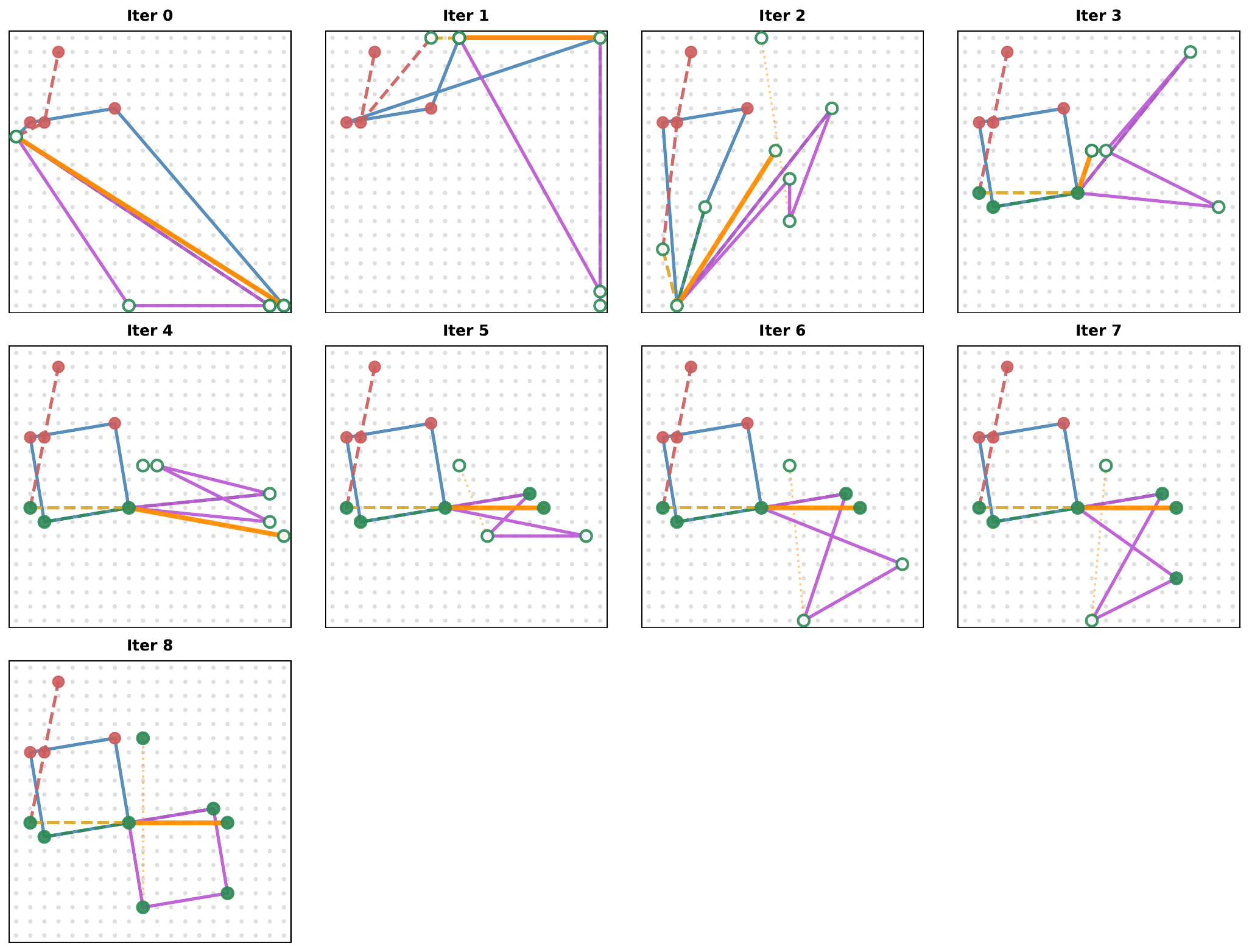}
    \caption{Problem with 6 constraints: two squares, three midpoints, and one reflection. One midpoint is less visible in the final solution because it is both a common vertex of both squares and simultaneously the midpoint of two different midpoint constraints. The orange line segment provides the reflection axis.}
    \label{fig:sol3}
\end{figure}

These examples cover three of the four constraint types in our CSP language: Midpoint ($M$), Reflection ($R$), and Square ($S$), providing broader evidence for the model's geometric reasoning capabilities. The instances shown in Figures~\ref{fig:sol1}, \ref{fig:sol2}, and \ref{fig:sol3} were selected primarily for visual clarity. Some generated problems place points in close proximity that obscures the iterative dynamics when visualized.

Unlike the prominent example in the main text, we omit point labels to avoid visual clutter while still clearly showing the constraint relationships through colored lines (for same reason only problems with low constraint amount and depth are presented). Each figure shows how the model progressively constructs the hidden geometric configuration, with constraint satisfaction improving over iterations until convergence to the correct solution. These examples reinforce our core finding that the GNN employs a continuous optimization-like process to solve geometric constraint problems, moving point embeddings iteratively toward configurations that satisfy the given constraints.

In these visualizations, known points appear as fixed red markers throughout all iterations, while unknown points are shown as green markers that move as the model iterates. Unknown points appear as empty circles when incorrectly positioned and filled circles when they reach their correct locations. We display all iterations until final resolution. Different constraint types use distinct visual representations: Square constraints appear as full lines connecting all four vertices, Midpoint constraints use dashed lines forming a three-point chain, and Reflection constraints show full lines connecting the axis points with less visible dotted lines connecting the reflected point pairs.

\section{Constraint Embeddings Analysis}\label{app:constraint_embeddings}

We analyze the information encoded in constraint embeddings produced by the GNN during its iterative process. These embeddings are updated using information from both fixed and unknown points.

\subsection{Constraint Type Classification}

\begin{figure}[h!t]
    \centering
    \includegraphics[width=0.8\textwidth]{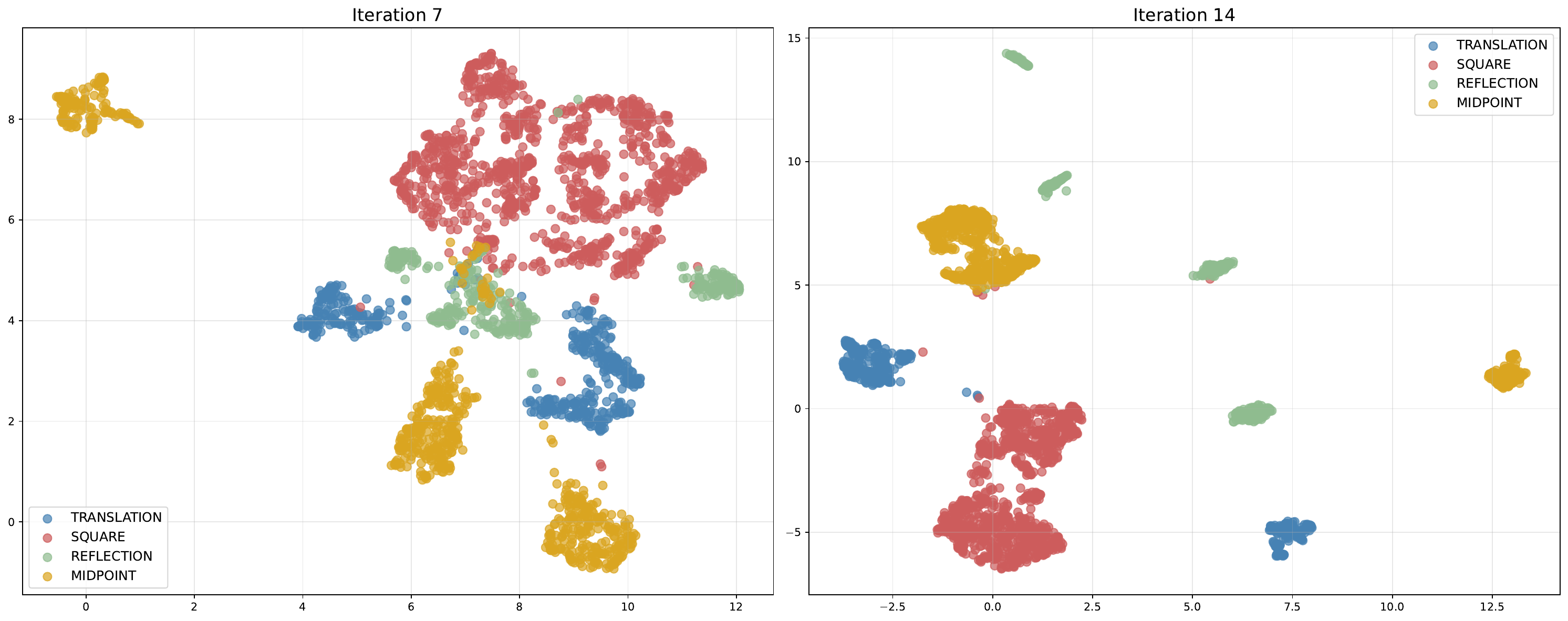}
    \caption{Constraint embedding clusters after $8$ and $15$ iterations, projected using UMAP. Different constraint types form distinct clusters, with subclusters reflecting geometric properties and network biases.}
    \label{fig:cstclusters}
\end{figure}

We tested whether constraint embeddings encode their constraint type by training a simple MLP classifier (two linear layers with ReLU) on the embeddings. The classifier achieved over $95~\%$ accuracy predicting constraint types. Figure~\ref{fig:cstclusters} shows the clear separation between constraint types in the projected embedding space using UMAP.

Each constraint type exhibits subclustering patterns that reflect geometric properties, network processing biases, and some generator design choices. Square constraints form subclusters based on side orientation relative to the grid (parallel versus diagonal orientations). Midpoint constraints cluster according to which variable is unknown: the midpoint itself or one of the endpoint variables. Reflection constraints show four subclusters corresponding to reflection axis orientation: two for axes parallel to grid edges and two for diagonal axes. Translation constraints exhibit order-dependent clustering based on which variables in the constraint $(A, B, C, D)$ are unknown---specifically whether variables $B, C$ or $A, D$ are unknown---revealing network bias toward variable ordering. Note that we identified fewer distinct subclusters than the maximum possible number, as some potential subclusters appeared very close in the embedding space.

These patterns indicate that constraint embeddings encode both geometric properties and structural biases from the network's processing order. As discussed in Section~\ref{sec:data_gen}, our generator creates problems requiring unique solutions through specific dependency structures, which may contribute to these ordering biases. Future work could explore making constraints invariant to variable permutations.

\subsection{Constraint Satisfaction Prediction}

We trained an MLP classifier to predict whether individual constraints are satisfied at each iteration. Using our $30$k test dataset, we annotated constraint satisfaction status after each iteration up to $15$ iterations and split the data: $70~\%$ for training, $10~\%$ for validation, and $20~\%$ for testing. We made sure that training data had balanced classes.

\begin{figure*}[ht]
  \centering
    \subfloat[]{\includegraphics[width=0.48\textwidth]{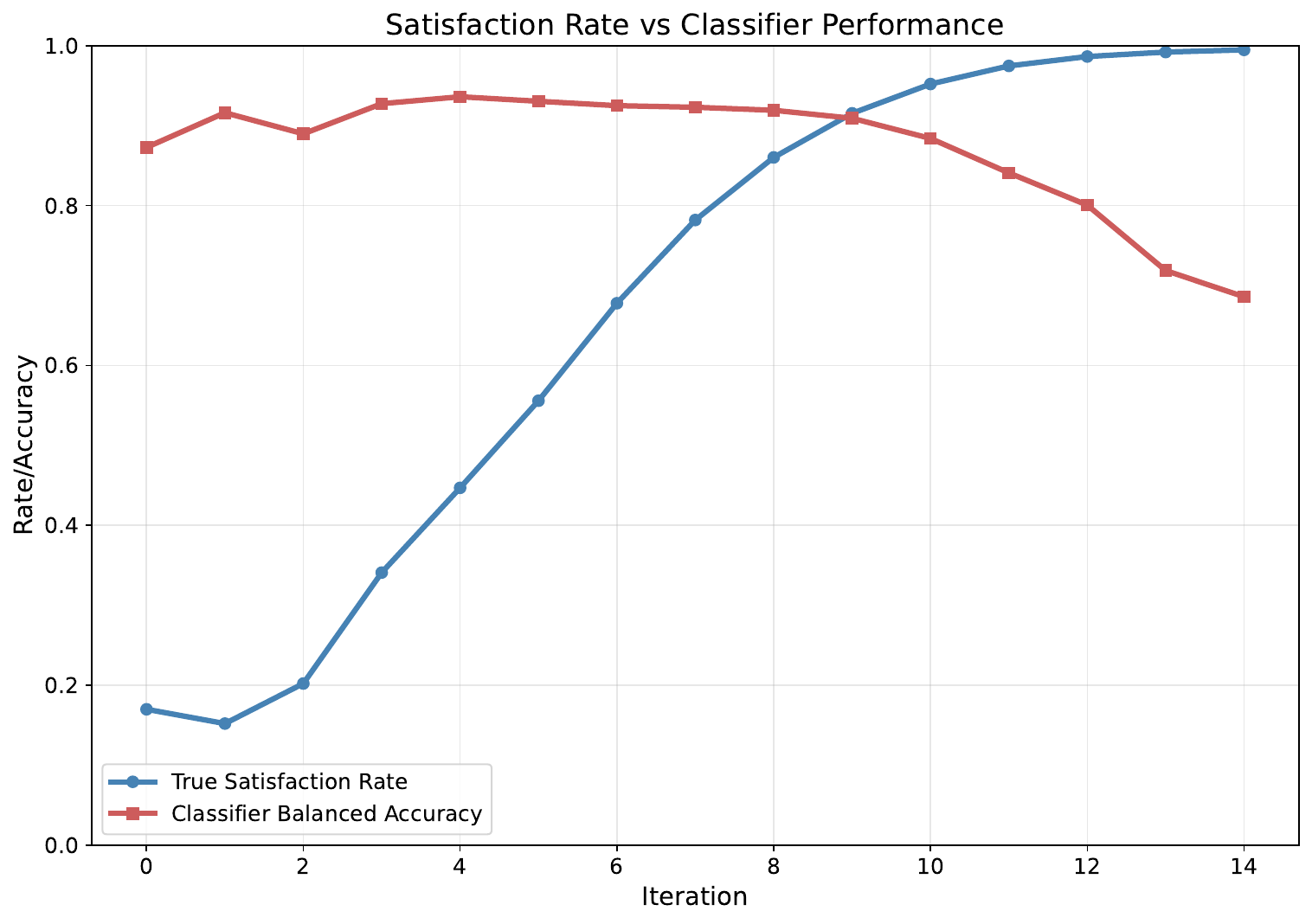}}
    \subfloat[]{\includegraphics[width=0.48\textwidth]{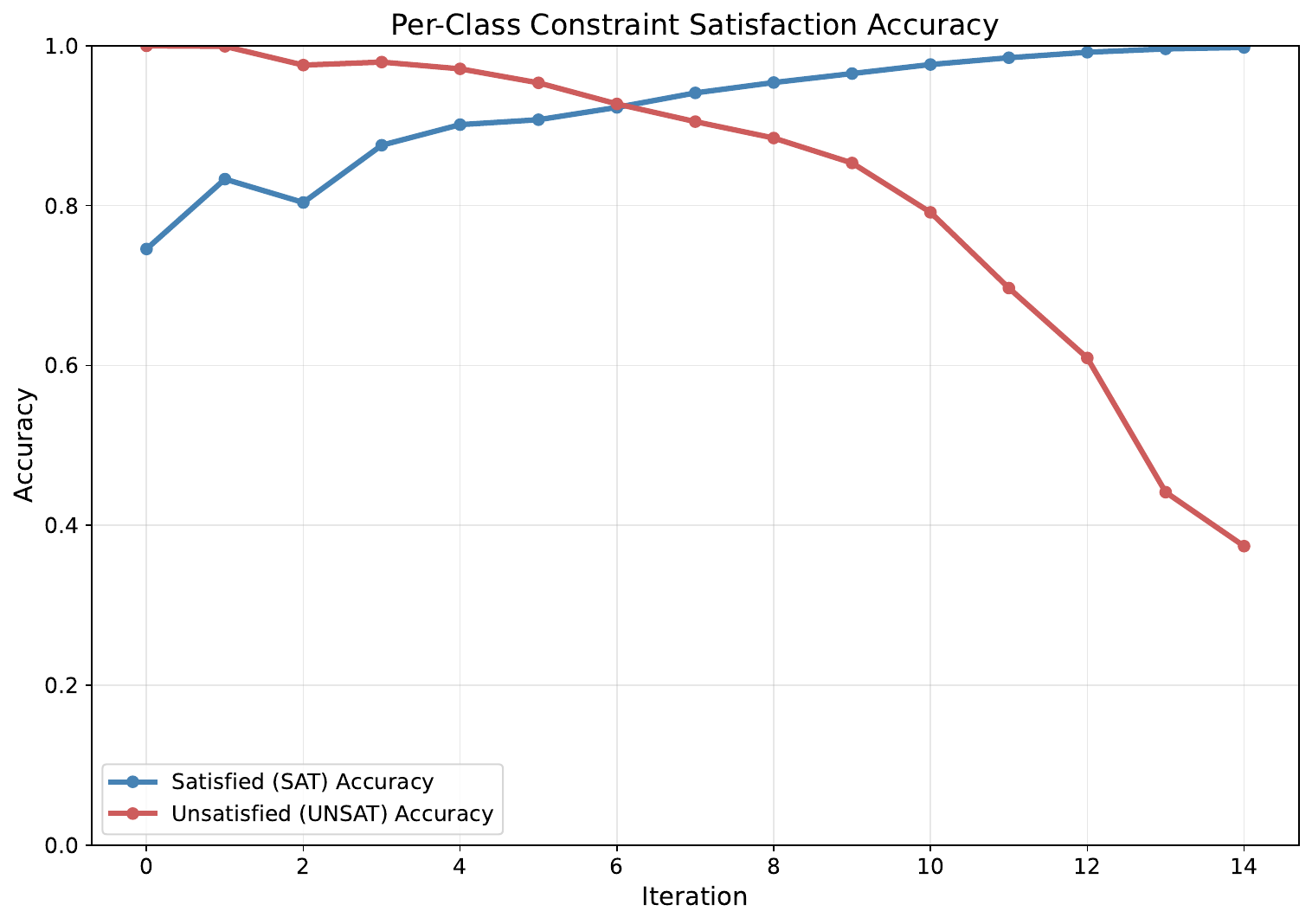}}
  \caption{Constraint satisfaction prediction accuracy across iterations. (a) True satisfaction rate versus classifier balanced accuracy. (b) Per-class accuracy showing the model's bias toward predicting ``satisfied'' at higher iterations.}\label{fig:cst_cls_acc}
\end{figure*}

For early iterations, the classifier achieved over $80~\%$ accuracy. However, performance degrades for higher iterations, as shown in Figure~\ref{fig:cst_cls_acc}. The model increasingly predicts ``satisfied'' for most constraints as iteration count increases, regardless of actual satisfaction status.

\subsection{Temporal Information Encoding}

\begin{figure}[h!t]
    \centering
    \includegraphics[width=0.65\textwidth]{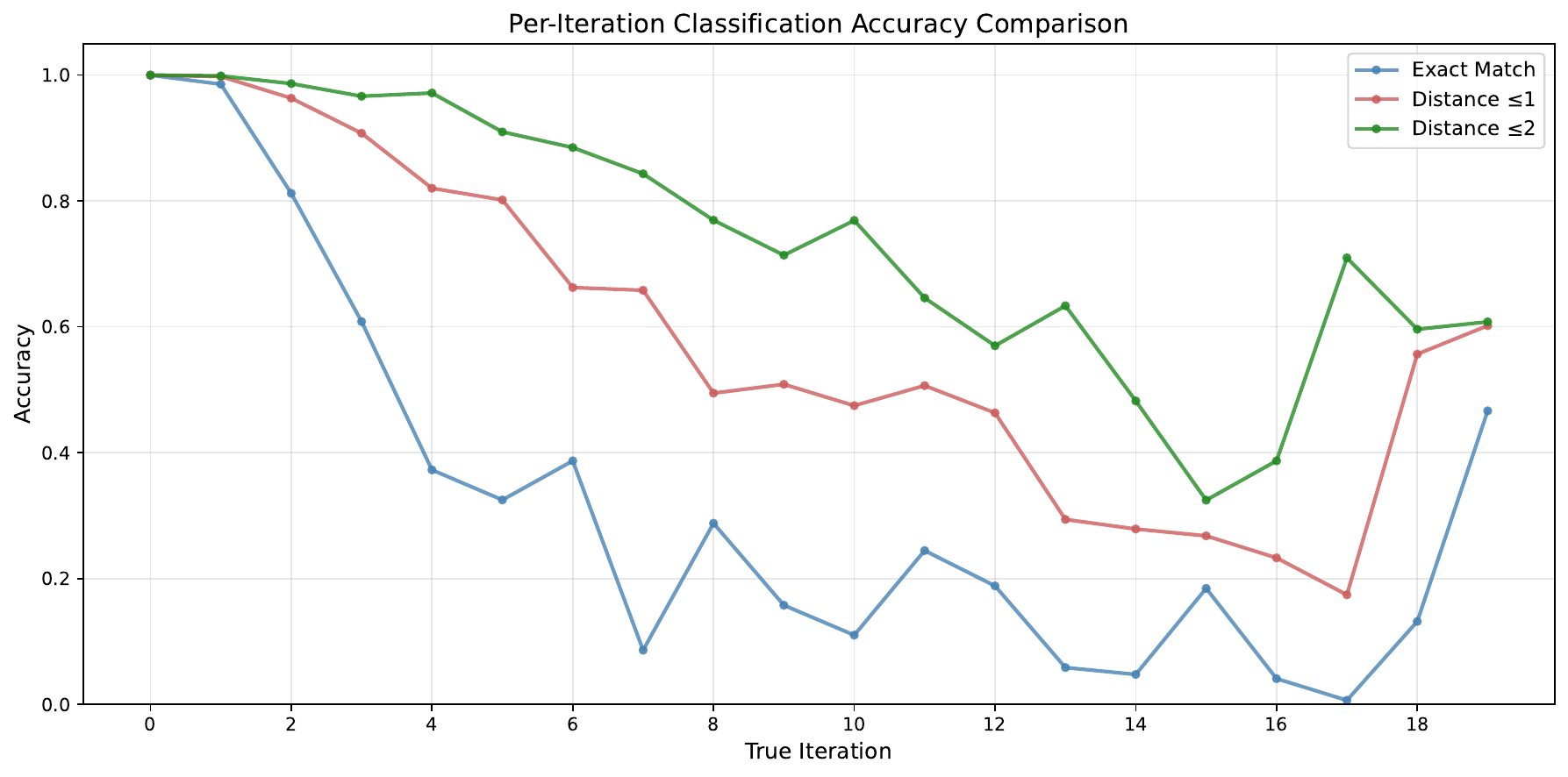}
    \caption{Accuracy of predicting iteration number from constraint embeddings. Exact match accuracy (blue), allowing distance $1$ (orange), and distance $2$ (green) tolerance. The model accurately predicts early iterations ($\leq 4$) but becomes less reliable for higher iteration counts.}
    \label{fig:cst_it_acc}
\end{figure}

We investigated whether constraint embeddings encode iteration number by training an MLP to predict the current iteration from constraint embeddings. Results using $20$ iterations are shown in Figure~\ref{fig:cst_it_acc}.

The classifier achieves high accuracy for early iterations ($\leq 4$) but becomes less reliable for higher iteration counts. When allowing tolerance (predicting within $1$-$2$ iterations of the true value), accuracy clearly improves. This indicates that constraint embeddings do encode temporal information, though with decreasing precision for later iterations.

When trained on higher maximum iteration counts, the predictor defaults to the final iteration class for later iterations, suggesting it learns a coarse ‘early’ vs ‘late’ distinction rather than precise temporal positioning. This may be due to the original GNN not being exposed to longer sequences during training.

These findings demonstrate that constraint embeddings encode rich information about constraint types, geometric relationships, satisfaction status, and temporal progression, providing more insight into the model's internal reasoning process.

\section{Evolution of Constraints under UMAP Projection}\label{sec:umap-evolution}

Unlike earlier visualizations which involved interpretation of classification outputs in a precise 2D grid, this analysis focuses on how the \emph{embedding space} itself organizes geometric reasoning. We use UMAP to project the point embeddings, both the static embeddings from the shared embedding layer and the evolving embeddings of unknown points during inference into 2D space. These projections are not tied to the output logits or grid structure, but purely reflect how the network shapes its internal representation geometry.

To visualize the reasoning dynamics, we perform the projection independently at each inference iteration. We then connect the points that participate in the same constraint using the same logic as in earlier visualizations in the precise 2D grid (i.e., if there is a line segment between points $A$ and $B$ in the shape for the given constraint, we add the segment to the projected image). This allows us to track how spatial relationships emerge and evolve directly in embedding space, without reference to grid positions or decoded predictions.

\begin{figure}[ht]
    \centering
    \includegraphics[width=0.85\textwidth]{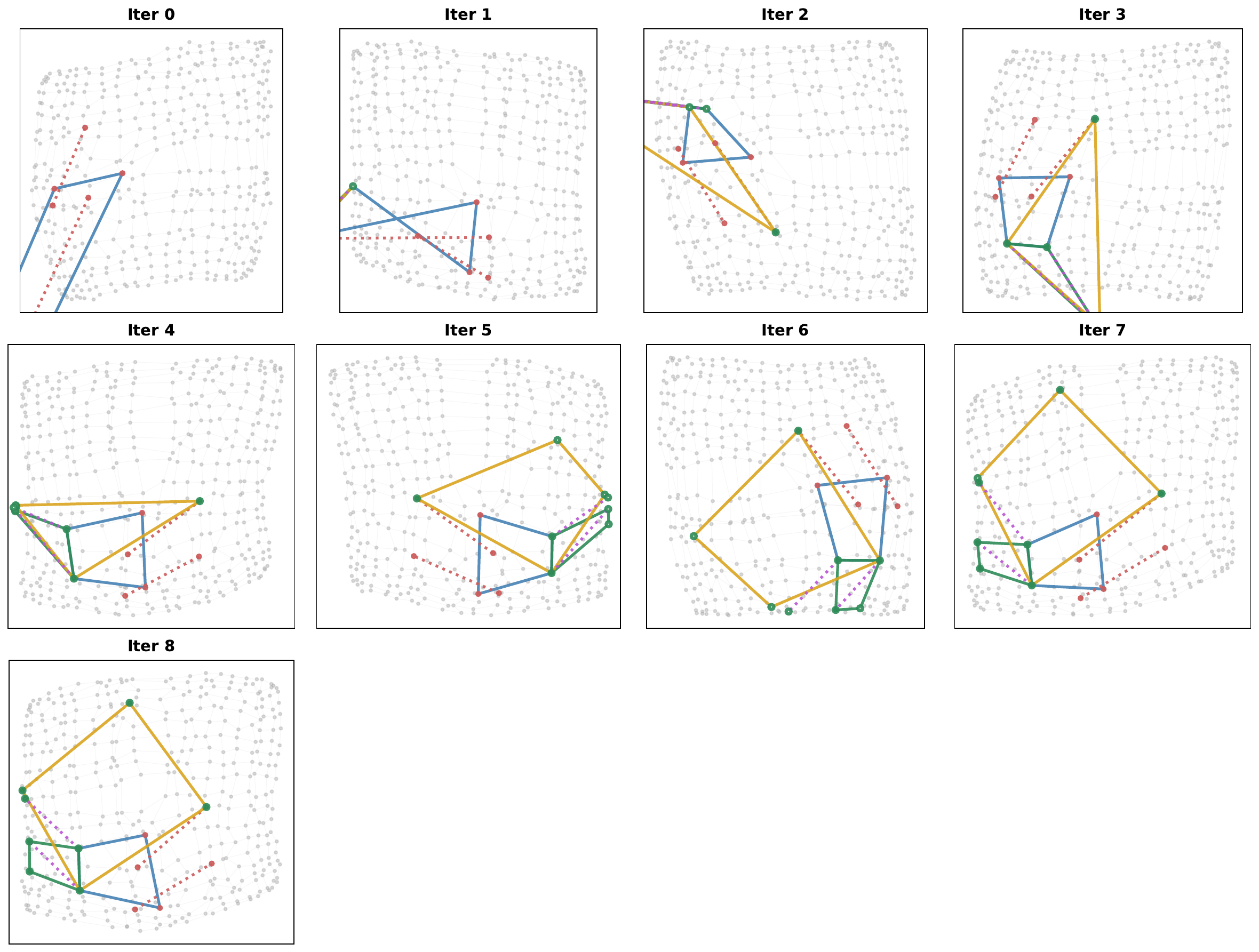}
    \caption{Constraint-based visualization of the UMAP projection of point embeddings across inference iterations for a simple problem. Each panel shows the embedding space at one iteration. Points are connected based on constraint structure. Colors denote known (red) and unknown (green) variables.}
    \label{fig:umapevo1}
\end{figure}

\begin{figure}[ht]
    \centering
    \includegraphics[width=0.85\textwidth]{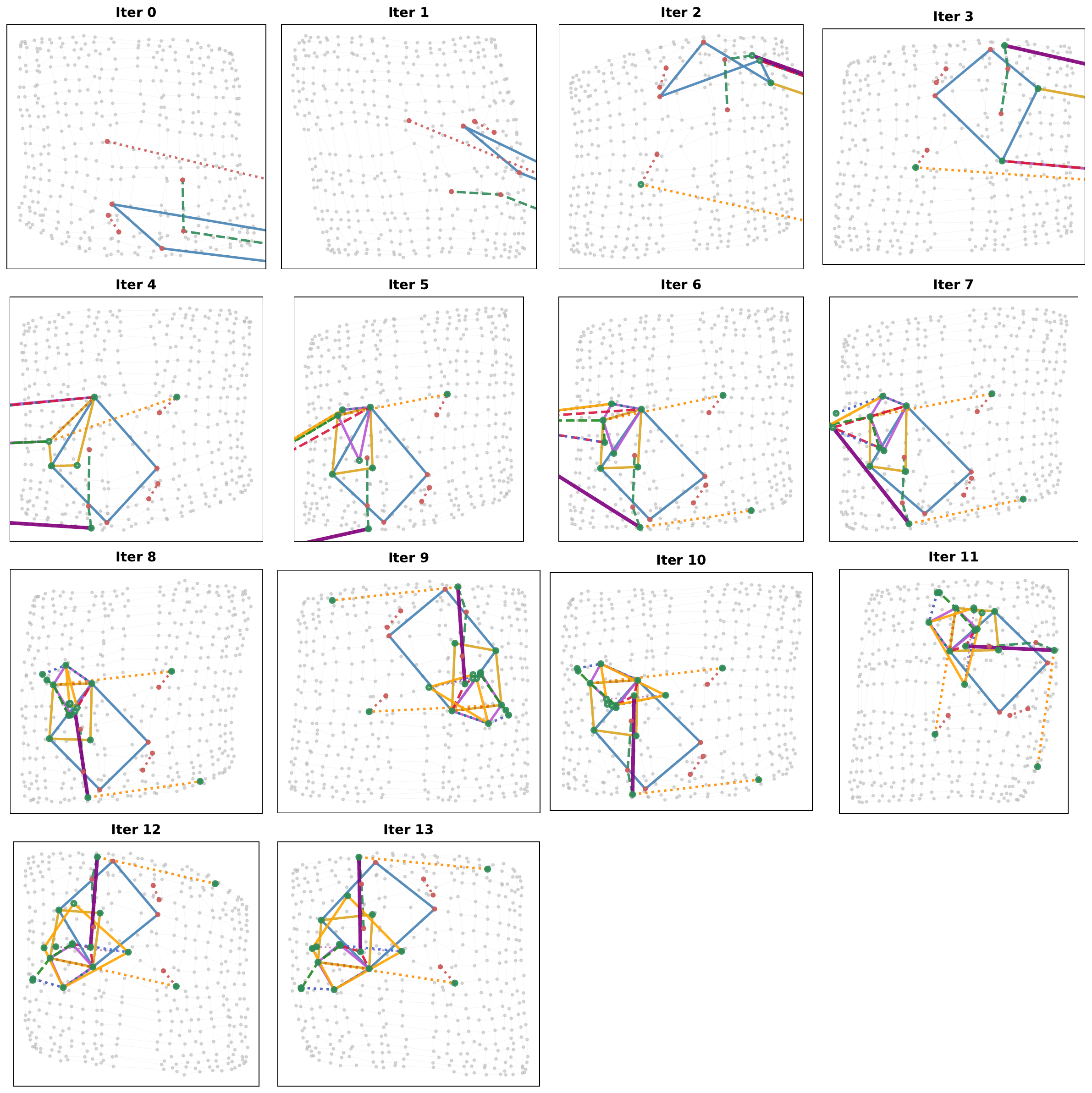}
    \caption{UMAP projections over inference time for a more complex instance with 13 constraints. Despite projection being performed independently per iteration, the geometric structure of the figure consistently emerges in embedding space. Constraint-based connections reveal progressive organization of the unknown points.}
    \label{fig:umapevo2}
\end{figure}

At the start of inference, the embeddings of unknown points are randomly initialized and lie far from the structured region formed by the fixed (grid) embeddings. During inference, especially within the first 5 iterations, these points are pulled into the manifold shaped by the fixed points of the grid, forming a coherent geometric configuration. This initial phase is clearly visible in the UMAP projections, reflecting the alignment of unknown embeddings with the learned spatial structure.

We also tried to quantify these dynamics by computing several geometric metrics over the UMAP projections at each iteration. We compute a set of 9 normalized metrics, each producing a score in $[0, 1]$, where $1.0$ indicates perfect satisfaction of a geometric property. These are grouped by constraint type:

\begin{itemize}
    \item \textbf{Square constraints:}
    \begin{itemize}
        \item \emph{Area ratio}: Compares the area of the quadrilateral to that of a square with the same average side length. Normalized to 1.0 when areas match.
        \item \emph{Side uniformity}: Measures the coefficient of variation of the four side lengths. A perfect square has all sides equal $\Rightarrow$ uniformity $= 1.0$.
        \item \emph{Corner angle quality}: Average deviation from $90^\circ$ angles at each corner, normalized so that perfect right angles give 1.0.
    \end{itemize}

    \item \textbf{Parallel / Equisegment constraints:}
    \begin{itemize}
        \item \emph{Length ratio}: Ratio of the two segment lengths, normalized as $\min(l_1, l_2) / \max(l_1, l_2)$. Equals 1.0 if both segments are the same length.
        \item \emph{Parallelism}: Absolute cosine similarity of the direction vectors of the two segments. Perfect alignment (parallel or antiparallel) yields 1.0.
    \end{itemize}

    \item \textbf{Diamond constraints:}
    \begin{itemize}
        \item \emph{Reflection accuracy}: Measures how close one point is to the mirror image of another across a specified axis. Defined via normalized distance to the ideal reflection point.
        \item \emph{Axis quality}: Assesses whether the axis bisects the reflected segment perpendicularly, computed as normalized distance from its midpoint to the axis.
    \end{itemize}

    \item \textbf{Midpoint constraints:}
    \begin{itemize}
        \item \emph{Collinearity}: Computed as $1 -$ normalized triangle area. Perfect alignment yields 1.0.
        \item \emph{Midpoint accuracy}: The best of three permutations, measuring how close one point is to the midpoint of the other two.
    \end{itemize}
\end{itemize}

\begin{figure}[ht]
    \centering
    \includegraphics[width=0.85\textwidth]{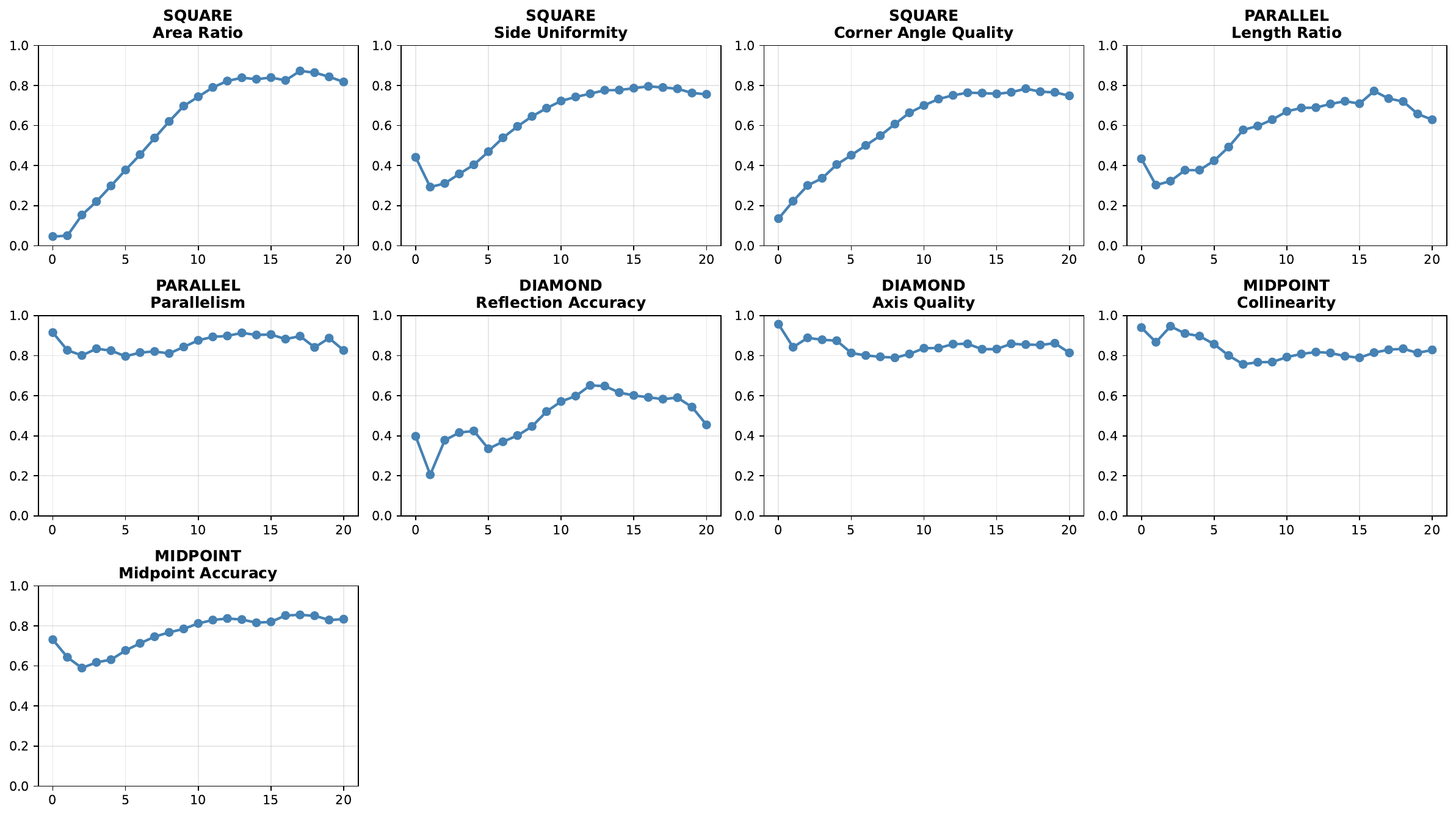}
    \caption{Evolution of geometric metrics computed on UMAP projections of point embeddings. Left: parallelism (axis alignment) across similar constraint types. Middle: average per-point movement across iterations (embedding consistency). Right: directional coherence for constraints of the same type (direction quality). All metrics are averaged across test instances and grouped by constraint type.}
    \label{fig:emapmetrics}
\end{figure}

Figure~\ref{fig:emapmetrics} shows the average value of these metrics per constraint type and iteration. The results indicate that geometric regularities emerge over time in the embedding space (even though no supervision is provided on these properties). This supports the interpretation that the network builds and organizes internal geometry as part of its reasoning process, and that UMAP is suited to uncover this structure (it would not work in original dimension or under PCA).

%% file: references.bib
@inproceedings{de2008z3,
  title={Z3: An efficient SMT solver},
  author={De Moura, Leonardo and Bj{\o}rner, Nikolaj},
  booktitle={International conference on Tools and Algorithms for the Construction and Analysis of Systems},
  pages={337--340},
  year={2008},
  organization={Springer}
}

@article{su2024roformer,
  title={Roformer: Enhanced transformer with rotary position embedding},
  author={Su, Jianlin and Ahmed, Murtadha and Lu, Yu and Pan, Shengfeng and Bo, Wen and Liu, Yunfeng},
  journal={Neurocomputing},
  volume={568},
  pages={127063},
  year={2024},
  publisher={Elsevier}
}

@article{dehghani2018universal,
  title={Universal transformers},
  author={Dehghani, Mostafa and Gouws, Stephan and Vinyals, Oriol and Uszkoreit, Jakob and Kaiser, {\L}ukasz},
  journal={arXiv preprint arXiv:1807.03819},
  year={2018}
}

@inproceedings{abbe2023generalization,
  title={Generalization on the unseen, logic reasoning and degree curriculum},
  author={Abbe, Emmanuel and Bengio, Samy and Lotfi, Aryo and Rizk, Kevin},
  booktitle={International Conference on Machine Learning},
  pages={31--60},
  year={2023},
  organization={PMLR}
}

@article{Wong2023FromWM,
  title={From Word Models to World Models: Translating from Natural Language to the Probabilistic Language of Thought},
  author={Li Siang Wong and Gabriel Grand and Alexander K. Lew and Noah D. Goodman and Vikash K. Mansinghka and Jacob Andreas and Joshua B. Tenenbaum},
  journal={ArXiv},
  year={2023},
  volume={abs/2306.12672},
  url={https://api.semanticscholar.org/CorpusID:259224900}
}

@article{hochreiter1997long,
  title={Long Short-term Memory},
  author={Hochreiter, S},
  journal={Neural Computation MIT-Press},
  year={1997}
}

@article{Ivanitskiy2023StructuredWR,
  title={Structured World Representations in Maze-Solving Transformers},
  author={Michael I. Ivanitskiy and Alex F Spies and Tilman Rauker and Guillaume Corlouer and Chris Mathwin and Lucia Quirke and Can Rager and Rusheb Shah and Dan Valentine and Cecilia G. Diniz Behn and Katsumi Inoue and Samy Wu Fung},
  journal={ArXiv},
  year={2023},
  volume={abs/2312.02566},
  url={https://api.semanticscholar.org/CorpusID:265659365}
}

@article{Yamada2023EvaluatingSU,
  title={Evaluating Spatial Understanding of Large Language Models},
  author={Yutaro Yamada and Yihan Bao and Andrew Kyle Lampinen and Jungo Kasai and Ilker Yildirim},
  journal={ArXiv},
  year={2023},
  volume={abs/2310.14540},
  url={https://api.semanticscholar.org/CorpusID:264426397}
}

@article{Momennejad2023EvaluatingCM,
  title={Evaluating cognitive maps and planning in large language models with cogeval},
  author={Momennejad, Ida and Hasanbeig, Hosein and Vieira Frujeri, Felipe and Sharma, Hiteshi and Jojic, Nebojsa and Palangi, Hamid and Ness, Robert and Larson, Jonathan},
  journal={Advances in Neural Information Processing Systems},
  volume={36},
  pages={69736--69751},
  year={2023}
}

@incollection{wu2008decision,
  title={On the decision problem and the mechanization of theorem-proving in elementary geometry},
  author={Wu, Wen-ts{\"u}n},
  booktitle={Selected Works Of Wen-Tsun Wu},
  pages={117--138},
  year={2008},
  publisher={World Scientific}
}

@article{wen1986basic,
  title={Basic principles of mechanical theorem proving in elementary geometries},
  author={Wen-Tsun, Wu},
  journal={Journal of automated Reasoning},
  volume={2},
  pages={221--252},
  year={1986},
  publisher={Springer}
}

@article{trinh2024solving,
  title={Solving olympiad geometry without human demonstrations},
  author={Trinh, Trieu H and Wu, Yuhuai and Le, Quoc V and He, He and Luong, Thang},
  journal={Nature},
  volume={625},
  number={7995},
  pages={476--482},
  year={2024},
  publisher={Nature Publishing Group UK London}
}

@inproceedings{krueger2021automatically,
  title={Automatically Building Diagrams for Olympiad Geometry Problems.},
  author={Krueger, Ryan and Han, Jesse Michael and Selsam, Daniel},
  booktitle={CADE},
  pages={577--588},
  year={2021}
}

@inproceedings{hula2024revisiting,
  title={Understanding GNNs for Boolean Satisfiability through Approximation Algorithms},
  author={H{\r{u}}la, Jan and Moj{\v{z}}{\'\i}{\v{s}}ek, David and Janota, Mikol{\'a}{\v{s}}},
  booktitle={Proceedings of the 33rd ACM International Conference on Information and Knowledge Management},
  pages={953--961},
  year={2024}
}

@inproceedings{wu2024mind,
  title={Mind's Eye of LLMs: Visualization-of-Thought Elicits Spatial Reasoning in Large Language Models},
  author={Wu, Wenshan and Mao, Shaoguang and Zhang, Yadong and Xia, Yan and Dong, Li and Cui, Lei and Wei, Furu},
  booktitle={The Thirty-eighth Annual Conference on Neural Information Processing Systems},
  year={2024}
}

@article{teodorescu2022spatialsim,
  title={SpatialSim: Recognizing Spatial Configurations of Objects with Graph Neural Networks},
  author={Teodorescu, Laetitia and Hofmann, Katja and Oudeyer, Pierre-Yves},
  journal={Frontiers in Artificial Intelligence},
  volume={4},
  pages={782081},
  year={2022},
  publisher={Frontiers Media SA}
}

@article{janner2018representation,
  title={Representation learning for grounded spatial reasoning},
  author={Janner, Michael and Narasimhan, Karthik and Barzilay, Regina},
  journal={Transactions of the Association for Computational Linguistics},
  volume={6},
  pages={49--61},
  year={2018},
  publisher={MIT Press One Rogers Street, Cambridge, MA 02142-1209, USA journals-info~…}
}

@inproceedings{li2023depwignn,
  title={Dep{W}i{GNN}: A Depth-wise Graph Neural Network for Multi-hop Spatial Reasoning in Text},
  author={Li, Shuaiyi and Deng, Yang and Lam, Wai},
  booktitle={Findings of the Association for Computational Linguistics: EMNLP 2023}, 
  year={2023}
}

@article{radford2019language,
  title={Language models are unsupervised multitask learners},
  author={Radford, Alec and Wu, Jeffrey and Child, Rewon and Luan, David and Amodei, Dario and Sutskever, Ilya and others},
  journal={OpenAI blog},
  volume={1},
  number={8},
  pages={9},
  year={2019}
}

@article{touvron2023llama,
  title={Llama: Open and efficient foundation language models},
  author={Touvron, Hugo and Lavril, Thibaut and Izacard, Gautier and Martinet, Xavier and Lachaux, Marie-Anne and Lacroix, Timoth{\'e}e and Rozi{\`e}re, Baptiste and Goyal, Naman and Hambro, Eric and Azhar, Faisal and others},
  journal={arXiv preprint arXiv:2302.13971},
  year={2023}
}

@article{wei2022chain,
  title={Chain-of-thought prompting elicits reasoning in large language models},
  author={Wei, Jason and Wang, Xuezhi and Schuurmans, Dale and Bosma, Maarten and Xia, Fei and Chi, Ed and Le, Quoc V and Zhou, Denny and others},
  journal={Advances in neural information processing systems},
  volume={35},
  pages={24824--24837},
  year={2022}
}

@article{selsam2018learning,
  title={Learning a SAT solver from single-bit supervision},
  author={Selsam, Daniel and Lamm, Matthew and B{\"u}nz, Benedikt and Liang, Percy and de Moura, Leonardo and Dill, David L},
  journal={arXiv preprint arXiv:1802.03685},
  year={2018}
}

@article{mojvzivsek2025neural,
  title={Neural Approaches to SAT Solving: Design Choices and Interpretability},
  author={Moj{\v{z}}{\'\i}{\v{s}}ek, David and H\r{u}la, Jan and Li, Ziwei and Zhou, Ziyu and Janota, Mikol{\'a}{\v{s}}},
  journal={arXiv preprint arXiv:2504.01173},
  year={2025}
}

@article{wong2022euclidnet,
  title={EuclidNet: Deep visual reasoning for constructible problems in geometry},
  author={Wong, Man Fai and Qi, Xintong and Tan, Chee Wei},
  journal={arXiv preprint arXiv:2301.13007},
  year={2022}
}
